\documentclass{article}

\usepackage{microtype}
\usepackage{graphicx}
\usepackage{subfigure}
\usepackage{booktabs} % for professional tables

\usepackage{hyperref}
\usepackage{tcolorbox}

\usepackage[accepted]{icml2025}
\usepackage{amsmath}

\usepackage{amssymb}
\usepackage{mathtools}
\usepackage{amsthm}
\usepackage{enumitem}
\usepackage[capitalize,noabbrev]{cleveref}

\theoremstyle{plain}

\theoremstyle{definition}

\theoremstyle{remark}

\usepackage[textsize=tiny]{todonotes}

\usepackage{xcolor}

\newcommand{\ignore}[1]{}
\newcommand{\names}{\textsc{HARPO}}
\newcommand{\namesm}{\textsc{OmniSapiens-7B 2.0}}
\newcommand{\namel}{\textsc{Heterogeneity-Aware Relative Policy Optimization}}

\usepackage{listings}
\lstdefinestyle{promptstyle}{
  basicstyle=\ttfamily\small,
  frame=single,
  columns=fullflexible,
  breaklines=true,
  showstringspaces=false
}

\icmltitlerunning{OmniSapiens: A Foundation Model for Social Behavior Processing via Heterogeneity-Aware Relative Policy Optimization}

\begin{document}

\twocolumn[
\icmltitle{OmniSapiens: A Foundation Model for Social Behavior Processing\\via Heterogeneity-Aware Relative Policy Optimization}




\begin{icmlauthorlist}
\icmlauthor{Keane Ong}{nus,mit}
\icmlauthor{Sabri Boughorbel}{ksa1}
\icmlauthor{Luwei Xiao}{nus}
\icmlauthor{Chanakya Ekbote}{mit}
\icmlauthor{Wei Dai}{mit}
\icmlauthor{Ao Qu}{mit}
\icmlauthor{Jingyao Wu}{mit}
\icmlauthor{Rui Mao}{ntu}
\icmlauthor{Ehsan Hoque}{ksa2}
\icmlauthor{Erik Cambria}{ntu}
\icmlauthor{Gianmarco Mengaldo\textsuperscript{$\dagger$}}{nus}
\icmlauthor{Paul Pu Liang\textsuperscript{$\dagger$}}{mit}
\end{icmlauthorlist}

\icmlaffiliation{mit}{Massachusetts Institute of Technology}
\icmlaffiliation{nus}{National University of Singapore}
\icmlaffiliation{ntu}{Nanyang Technological University}
\icmlaffiliation{ksa1}{Prince Sattam bin Abdulaziz University}
\icmlaffiliation{ksa2}{University of Rochester}

\icmlcorrespondingauthor{Keane Ong}{keaneong@mit.edu}

\icmlkeywords{Machine Learning, ICML}

\vskip 0.3in
]


\printAffiliationsAndNotice{\textsuperscript{$\dagger$}Equal advising.}

\begin{abstract}
Socially intelligent AI systems must reason across diverse human behavioral tasks and generalize to new social contexts. However, behavioral data is inherently heterogeneous, comprising diverse modalities and prediction targets that produce uneven training signals across samples, creating imbalanced learning dynamics that challenge existing AI models. To address this, we develop \namesm, a foundation model for social behavior processing that explicitly addresses learning from heterogeneous behavioral data. This is enabled through \namel, a new RL method that rebalances learning signals across samples by approximating each sample's contribution to the policy update and using these estimates to drive geometrically centered, inertially smoothed advantage modulation for stable training. \namesm\ achieves the best and most consistent performance across 10 behavioral tasks, while also attaining the best performance on all five held-out benchmarks, with gains of up to +12.02\% and +9.37\% respectively. Furthermore, it demonstrates more consistent and interpretable reasoning traces, supporting reliable real-world behavioral applications. Our model is available at \url{https://github.com/MIT-MI/human_behavior_atlas}.

 % that explicitly rebalances learning signals across samples
\end{abstract}

% \paul{the technical novelty is not highlighted sufficiently in this 1 sentence, what was challenging and what was the key insight?}
 % that achieve state-of-the-art performance by designed to learn effectively across heterogeneous behavioral signals

\vspace{-6mm}
\section{Introduction}\label{sec:intro}

Social intelligence, the ability to interpret affective expressions, mental states, and social signals, does not emerge from an isolated understanding of behavioral cues~\cite{baron2001intercontheory, goleman2006social}. 
How can we infer intent without perceiving social context? 
How can we understand sarcasm without understanding humor? 
Social intelligence requires reasoning across interconnected behavioral contexts, where distinct signals interact to produce coherent understanding and adaptation to new situations~\cite{kihlstrom2000social}. Yet, existing AI systems are predominantly specialized for a single behavioral task (e.g., sentiment analysis), with limited progress toward unified models that can effectively generalize across diverse behavioral contexts.

% Yet existing AI systems remain largely fragmented into narrow task-specific models, and there still lacks a unified foundation model capable of jointly reasoning across diverse behavioral phenomena. This fragmentation arises in part from the intrinsic heterogeneity of behavioral data, where diverse modalities, objectives, and supervision structures induce highly uneven optimization dynamics during training.
 % To start with, behavioral signals are inherently ambiguous and complex, often benefiting from reasoning capabilities~\citep{scherer2009dynamic}.
% Yet, current AI systems remain largely specialized for narrowly defined behavioral understanding tasks
% \keane{see if can be neater}
To develop unified social behavioral models, recent work has curated large-scale behavioral benchmarks~\cite{ong2025human} and adopted reasoning-based RL~\cite{zhang2025surveyreasoning} to elicit the structured reasoning behavioral tasks require. 
However, unified models still remain significantly constrained by the imbalanced learning dynamics induced from behavioral data. 
Behavioral data is highly heterogeneous, spanning distinct feature types (e.g., acoustic prosody, interpersonal interaction) and prediction targets (e.g., emotion classification~\citep{zadeh2016mosi}, social inference~\citep{siq2}), which often produces uneven training signals across samples. 
Existing unified models~\cite{ong2025human} typically employ RL methods, i.e., GRPO~\citep{shao2024deepseekmath}, to enhance reasoning performance. However, these methods aggregate policy updates across heterogeneous samples without explicitly regulating disparities in their learning dynamics. 
As a result, optimization can become disproportionately dominated by specific behavioral tasks or samples, hindering multitask performance and cross-context generalization.

To address these challenges, we build \namesm, a foundation model for social behavior processing across 10 behavioral tasks. 
\namesm\ is trained with \namel\ (\names), an on-policy optimization mechanism that rebalances learning across heterogeneous behavioral data to ensure that no single task or sample disproportionately influences policy optimization. This is achieved by approximating contribution signals to the policy update, and using them to inform geometrically centered and inertially smoothed advantage modulation.

Across 10 behavioral tasks spanning diverse cognitive, affective, pathological, and social phenomena, \namesm\ achieves significantly stronger multitask performance than existing behavioral models, with improvements of up to +12.02\%. Beyond multitask learning, \namesm\ further demonstrates the strongest zero-shot generalization to five held-out behavioral benchmarks spanning clinical behavior, affective expression, and social-pragmatic reasoning, outperforming prior models by up to +9.37\%. Beyond predictive performance, \namesm\ produces more interpretable reasoning traces, improving key XAI interpretability desiderata including robustness, efficiency, understandability, completeness, and compactness. Finally, compared against recent critic-free reasoning RL methods,  \names\ achieves the most consistently strong performance across behavioral tasks, outperforming GRPO by up to +42.29\%.

% Our key contribution is two-fold: (1) We develop \names, a novel critic-free, reasoning RL method for learning diverse behavioral tasks associated with heterogeneous data; (2) Leveraging \names, we train \namesm, a foundation model for unified human behavior
% analysis that substantially addresses the performance limitations of prior unified models, enabling effective performance across 10 diverse human behavior tasks and generalization to novel behavioral settings. For future work, models and codes will be made publicly available after the review process. 

Our contributions are as follows: (1) We develop and release \namesm, a foundation model for unified social behavior analysis. \namesm\ is one of the first models to jointly tackle a wide range of behavioral analysis tasks spanning affective, cognitive, pathological, and social dimensions. \namesm\ significantly outperforms prior state-of-the-art models across 10 diverse behavioral tasks, while also demonstrating substantially stronger generalization in novel behavioral settings. Furthermore, \namesm\ produces more interpretable reasoning traces, improving transparency and reliability for real-world behavioral analysis applications. Together, these capabilities unlock potential applications in affective computing, behavioral health, and socially intelligent AI systems, particularly in settings where behavioral data is scarce and model interpretability is critical; (2) we address the challenges associated with learning from heterogeneous behavioral data by introducing \names, a critic-free, reasoning-based RL method designed to learn from heterogeneous behavioral data.

% \keane{novelty as an new allocator for critic-free training, wherein stability is more determined by advantage construction and normalization.}

\section{Related Work}

%\textbf{Models for Understanding Psychological and Social Behaviors} \textcolor{green}{Jingyao: added; citations will be added soon later}

%Psychological and social behavior modeling spans multiple domains of human behavior and is essential for building the next generation of socially intelligent systems. It encompasses the modeling of affective states (arousal, valence, emotions), cognitive states (stress, cognitive load), mental health conditions (depression, anxiety), and social processes (humor).

\textbf{Social behavior processing} develops AI for modelling human behavioral markers. Existing studies emphasize task-specific modeling, including interpreting \textit{affective states} via emotion and arousal–valence classification~\citep{zadeh2016mosi, dang2023constrained}; \textit{cognitive states} through stress and cognitive load estimation~\citep{giannakakis2019review}; \textit{pathological states} by detecting depression and anxiety~\citep{joshi2022depression, miloyan2014future}; and \textit{social processes} through humor and engagement detection~\citep{hessel2023androids,monkaresi2016automated}. While useful, task-specific processing overlooks the interdependencies between behavioral dimensions~\citep{pessoa2008relationship}, motivating unified approaches for representation sharing~\citep{ong2025human}.

% and affective state ranking~\citep{yannakakis2018ordinal, wu2022novel} akccay2020speech,
% Evidence from psychology and social science suggests that these dimensions are highly interrelated, with shared mechanisms shaping human behavior across contexts~\cite{pessoa2008relationship}. Yet, most existing works study these dimensions in isolation and lack of a unified behavioral modeling, overlooking their potential interrelations and the shared knowledge that could be exploited during model learning. 

%\textcolor{green}{jingyao: do we also emphasis on multimodal in this paper, if so, need to add discussion of unimodal \& multimodal modelling for human behaviour tasks.}
%studies how computational models understand, intepretate, predict and reason about

% \keane{insert table to compare between models} \textcolor{green}{jingyao: like the one in iclr paper?} \keane{yup}

% \textcolor{green}{Jingyao: @keane could you pick the points you think link better with our claims and connect them?} \keane{sure; maybe we should also add bits on GPG, RLOO}
\textbf{Reasoning-based reinforcement learning} has included recent advances such as GRPO~\citep{shao2024deepseekmath}  for group-normalized rewards, REINFORCE Leave-One-Out~\citep{ahmadian2024rloo} which reduces gradient variance, REINFORCE++~\citep{hu2025reinforce++} which improves optimization stability, and Group Policy Gradient~\citep{chu2025gpg} which models group-level objectives. Yet, reasoning RL for learning heterogeneous behavioral data remains underexplored.

% In contrast, GRPO-LEAD~\citep{zhang2025grpo_lead} focuses on shaping reasoning length and difficulty-aware reweighting.
% SofT-GRPO~\citep{zheng2025soft} and

% SofT-GRPO~\citep{zheng2025soft}: reinforce LLMs under the soft-thinking reasoning pattern. 
% GRPO-LEAD~\citep{zhang2025grpo_lead} in addressing the challenges in reward sparsity, verbosity, and inadequate focus on problem difficulty.

% i-MENTOR~\citep{gao2025navigate} to deliver dense rewards and amplify exploration in the RL-based paradigm.
% MO-GRPO~\citep{ichihara2025mo} focus on multi-objective settings that addresses the challenge when GRPO is vulnerable to reward hacking, optimizing only one of the objectives at the cost of the others. 
%~\citet{ma2024highly} introduce a self-adaptive and highly efficient reward shaping mechanism to address the sparse-reward problem in RL.
% Group Filtered Policy Optimization (GFPO)~\citep{shrivastava2025sample}: that filters training samples by response length and reward efficiency to mitigate reasoning length inflation in RL–trained language models.

%\textbf{Multitask Learning} \keane{try to look at: multiplicative weights, weighing via grad norm etc, but these algorithms somehow have not been applied for reasoning-RL algos}
\textbf{Multitask learning} has evolved from unimodal representation sharing to any-to-any architectures that jointly model multiple modalities~\citep{liang2024foundations}. Prior work has also explored gradient balancing~\citep{yu2020gradient}, uncertainty-based weighting~\citep{kendall2018multi}, shared distilled policies~\citep{teh2017distral}. Yet, multitask learning remains largely underexplored for recent reasoning RL.

\section{OmniSapiens-7B 2.0}\label{sec:full_model}

\subsection{Model Setup}

\textbf{Multimodal backbone.}  Given the multimodal nature of the behavioral data (text, vision, and audio)~\citep{liang2024foundations}, we adopt Qwen 2.5 Omni-7B~\citep{xu2025qwen2.5omni} as our base architecture for \namesm. All training follows a multitask setup, where a single model is jointly trained across all tasks.

\textbf{Behavioral training data.} We use the Human Behavior Atlas benchmark~\citep{ong2025human} for training, which covers diverse human behavior tasks over 100k samples. These include sentiment polarity (SEN), emotion recognition (EMO), social reasoning (SOC), intent recognition (INT), non-verbal communication (NVC), as well as detecting humor (HUM), sarcasm (SAR), anxiety (ANX), depression (DEP), and PTSD (PTSD). For training, we develop HARPO, a reasoning-RL algorithm that explicitly balances heterogeneous learning signals across behavioral tasks.

\subsection{Preliminaries}~\label{sec:prelim}
\vspace{-20pt}

\textbf{Problem: Learning multiple behavioral tasks.} 
We consider the problem of training a stochastic policy $\pi_\theta(a \mid s)$, instantiated as a large language model (LLM), to perform a diverse set of behavioral understanding tasks indexed by $m \in \mathcal{M}$. For each task $m$, a sample $q \sim \mathcal{D}_m$, which can include multimodal inputs (e.g., text, audio, or visual signals), is drawn from a task-specific input distribution, and the policy generates an autoregressive output sequence $o \sim \pi_\theta(\cdot \mid q)$, with tokens $o_{:k}$ and prefixes $o_{:<k}$.  The learning objective is to train a single shared policy $\pi_\theta$ that maximizes performance across all tasks $m \in \mathcal{M}$.

% Task-dependent reward functions $r_m(q, o)$ evaluate the quality of generated outputs. The learning objective is to maximize expected reward across tasks:
% \begin{equation}
% \max_\theta \;\; 
% \mathbb{E}_{m \sim \mathcal{M}} \,
% \mathbb{E}_{q \sim \mathcal{D}_m} \,
% \mathbb{E}_{o \sim \pi_\theta(\cdot \mid q)}
% \big[ r_m(q, o) \big].
% \end{equation}

% To optimize this objective in practice, we consider Group Relative Policy Optimization (GRPO), an on-policy, critic-free method that has shown strong performance in recent LLM training.

\textbf{Group Relative Policy Optimization for behavioral tasks.} To optimize  $\pi_\theta(a\mid s)$, we consider Group Relative Policy Optimization (GRPO)~\citep{shao2024deepseekmath}, a recent on-policy RL method that has shown strong performance for reasoning-based LLM training. For task $m$ and sample $q$, GRPO samples a rollout group $G_{(m,q)}$ of responses $\{o_{(m,q,i)}\}$, where $i \in G_{(m,q)}$ indexes individual rollouts (i.e., a sampled response) with rewards $r_{(m,q,i)}$, computing the group-normalized advantage:
\begin{equation}
\small
\label{eq:grpo_adv}
\hat A_{(m,q,i)}
=
\frac{r_{(m,q,i)}-\hat\mu_{G_{(m,q)}}}{\hat\sigma_{G_{(m,q)}}+\varepsilon}
\end{equation}
where $\hat\mu_{G_{(m,q)}}$ and $\hat\sigma_{G_{(m,q)}}$ are the mean
and standard deviation of $\{r_{(m,q,i)}\}_{i=1}^{|G_{(m,q)}|}$. With a PPO clipped surrogate \( \tilde A_{(m,q,i):k}(\theta) \) constructed from \( \hat A_{(m,q,i)} \), GRPO then optimizes $\pi_\theta(a\mid s)$ using a PPO-style trust-region objective\footnote{For completeness, the full formulation of GRPO is provided in App.~\ref{app:grpo_full_formulation}.}:
\begin{equation}
\small
\begin{aligned}
J_{\mathrm{GRPO}}(\theta)
&=
\mathbb{E}_{(m,q)\sim \mathcal D}
\mathbb{E}_{\{o_{(m,q,i)}\}\sim \pi_{\theta_{\mathrm{old}}}}\!\Bigg[
\frac{1}{|G_{(m,q)}|}
\sum_{i\in G_{(m,q)}}
\\
&\hspace{-60pt}\phantom{=\mathbb{E}\Bigg[}
\frac{1}{n_{o_{(m,q,i)}}}
\sum_{k=1}^{n_{o_{(m,q,i)}}}
\tilde A_{(m,q,i):k}(\theta)
\Bigg]
\;-\;
\beta\,
\mathbb{E}\!\left[
D_{\mathrm{KL}}\!\left(\pi_\theta\;\|\;\pi_{\mathrm{ref}}\right)
\right]
\end{aligned}
\end{equation}

By the policy-gradient theorem~\citep{sutton1999policy}, the policy gradient admits the standard form, where the expectation is over $(s,a)$ induced by $\pi_\theta$:
\begin{equation}
\small
\label{eq:pg_base}
\nabla_\theta J(\theta)
=
\mathbb{E}\!\left[
A^{\pi_\theta}(s,a)\,\nabla_\theta \log \pi_\theta(a\mid s)
\right]
\end{equation}

When GRPO optimizes a policy across diverse behavioral tasks, gradient
contributions from different $m$, $q$, and $i$ are aggregated into a shared update.
Accordingly, a Monte Carlo estimator of the policy gradient, $g(\theta)$, can be decomposed as:
\begin{equation}
\small
\label{eq:pg_decomposition}
g(\theta)
=
\sum_{m \in \mathcal{M}}
\sum_{q \sim \mathcal{D}_m}
\sum_{i \in G_{(m,q)}}
g_{(m,q,i)}(\theta)
\end{equation}

Each rollout contributes a gradient term of the form:
\begin{equation}
\small
\label{eq:pg_component}
g_{(m,q,i)}(\theta)
\triangleq
\hat A_{(m,q,i)}\,
\nabla_\theta \log \pi_\theta(a_{(m,q,i)}\mid s_{(m,q,i)})
\end{equation}

% The relative contribution of a rollout to the
% overall update can be approximated through its normalized gradient:
% \begin{equation}
% \label{eq:implicit_allocation}
% \mathcal{B}_{(m,q,i)}
% =
% \frac{\|g_{(m,q,i)}(\theta)\|^{2}}{\sum_{m',q',i'} \|g_{(m',q',i')}(\theta)\|^{2}}.
% \end{equation}

% Using Eq.~\eqref{eq:pg_component}, this can be expanded as:
% \begin{equation}
% \label{eq:implicit_allocation_expanded}
% \resizebox{\columnwidth}{!}{$
% \begin{aligned}
% \mathcal{B}_{(m,q,i)}
% =
% \frac{
% \hat A_{(m,q,i)}^{2}
% \left\|
% \nabla_\theta \log \pi_\theta(a_{(m,q,i)} \mid s_{(m,q,i)})
% \right\|^{2}
% }{
% \sum_{m',q',i'}
% \hat A_{(m',q',i')}^{2}
% \left\|
% \nabla_\theta \log \pi_\theta(a_{(m',q',i')} \mid s_{(m',q',i')})
% \right\|^{2}
% }
% \end{aligned}
% $}
% \end{equation}

From Eq.\eqref{eq:pg_component}, each rollout contributes a gradient term whose magnitude is scaled by its advantage value. Since these rollout-level gradients are aggregated to form the shared policy gradient in Eq.\eqref{eq:pg_decomposition}, the resulting policy update is sensitive to the scale of advantages across rollouts. Across diverse behavioral tasks $m$, reward and advantage distributions may vary considerably because the associated behavioral data are highly heterogeneous, spanning different multimodal features (e.g., acoustic prosody, facial expressions) and prediction targets (e.g., emotion classification, social inference). Consequently, rollouts from tasks or samples with systematically elevated or suppressed advantage magnitudes can exert disproportionate influence on GRPO's policy update, contributing to uneven learning across tasks (Sec.~\ref{sec:ablations} provides an empirical illustration; App.~\ref{app:additional_plots} Fig.~\ref{fig:appendix_adv_dist} shows the differences in task advantage distributions). 

\subsection{Heterogeneity-Aware Relative Policy Optimization}~\label{sec:harpo}
\vspace{-20pt}

% a novel RL method that dynamically regulates how heterogeneous components contribute to a shared policy update. 

% The objective is to perform online allocation of the policy update budget across competing components, to promote balanced policy optimization. In line with (I) and (II), \names\ modulates component contributions to the shared update by redistributing post-GRPO advantages, preventing dominant advantage regimes from monopolizing optimization. Consistent with (III), this redistribution is performed online as component contributions evolve stochastically from on-policy sampling.

\textbf{Dynamic advantage modulation mechanism.} To address this gap, we introduce \namel\ (\names). Since rollout influence on the policy update scales with advantage magnitude, Eq.~\eqref{eq:pg_decomposition}, \names's core insight is to modulate advantages such that no single task or sample disproportionately influences policy optimization. In practice, our modulation mechanism scales the GRPO group-normalized advantages, Eq.~\eqref{eq:grpo_adv}, to mitigate imbalances in advantages at two levels: the sample-level, corresponding to advantages in a sample's rollout group, and the task-level, corresponding to advantages across all rollout groups of a task. This scaling is updated dynamically across training, using estimates of the relative contribution of each sample and task to the policy update. Concretely, we first construct a contribution signal $p^{(t)}$ from advantages, which approximates each task’s and sample’s relative contribution to the shared policy update.  
These signals are then transformed into structured, geometrically centered modulation factors, $\{ s_{(m,q)}^{(t)},\; s_m^{(t)} \}$, that scale the advantages prior to the policy update step.  
Finally, the modulation factors are updated using inertial smoothing to ensure their stability.

\begin{figure*}[h]
  \centering
\includegraphics[width=\textwidth]{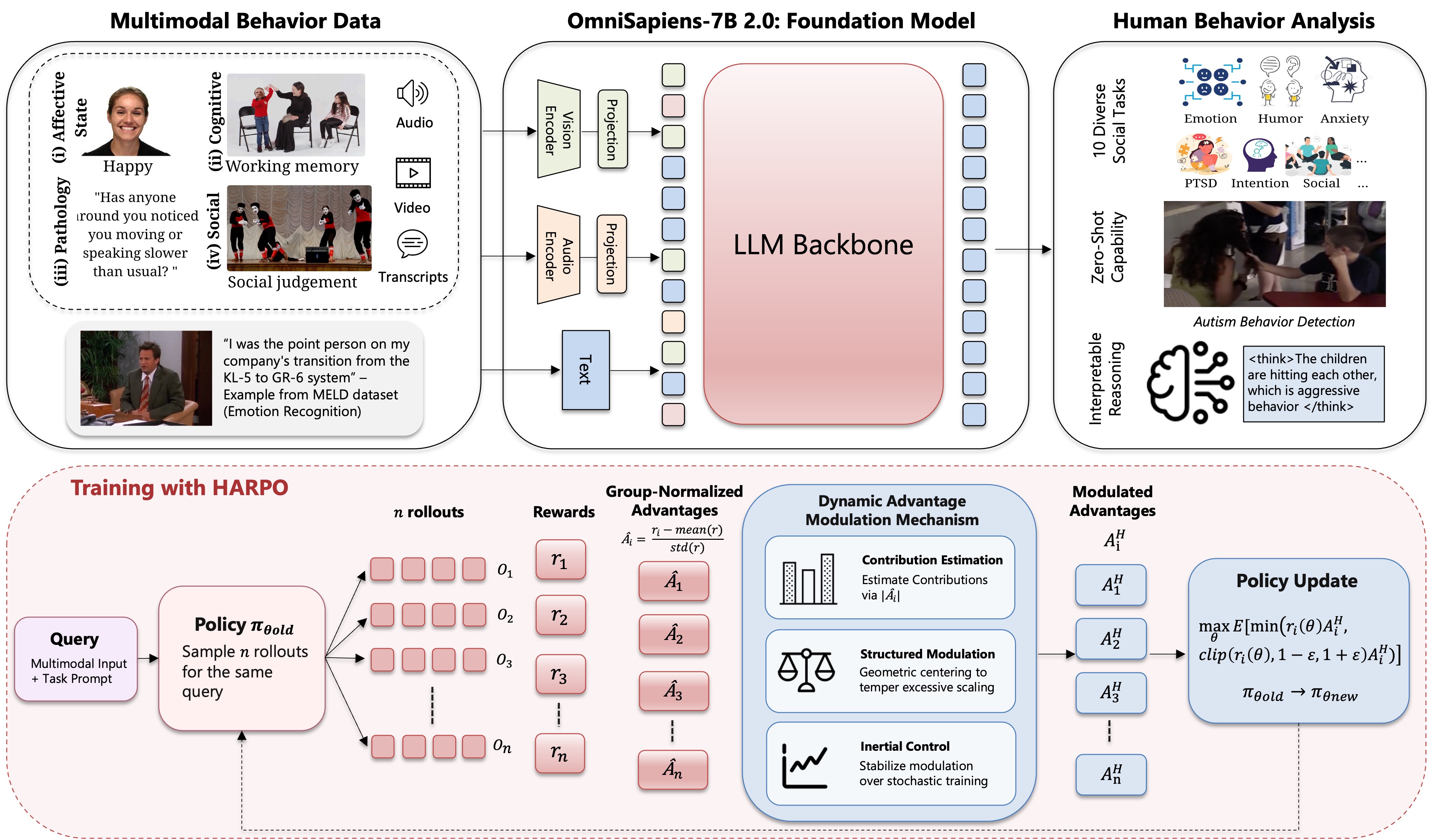}
\vspace{-20pt}
\caption{Model overview. \namesm\ significantly outperforms prior state-of-the-art models across 10 diverse behavioral tasks and demonstrates substantially stronger generalization in novel behavioral settings. In addition, \namesm\ produces relatively more interpretable reasoning traces, improving transparency and reliability for real-world behavioral analysis applications.}
\vspace{-12pt}
  \label{fig:omnisapiens_overview}
\end{figure*}

\textbf{Contribution signals from advantage magnitudes.}
% Absolute values are taken to measure contribution independent of sign, while taking the mean prevents $p$ from being dominated solely by sampling frequency. 
We construct a proxy signal, $p^{(t)}$, to approximate contribution strength to the policy update. Since policy gradients are directly scaled by advantages, Eq.\eqref{eq:pg_decomposition}-\eqref{eq:pg_component}, advantage magnitudes provide a computationally convenient proxy for contribution strength. Therefore, we use advantage magnitudes, normalized by rollout count to ensure invariance to stochastic batch sampling~\citep{schulman2017proximal}, and use this to define $p^{(t)}$ at two levels. While the sample-level $p_{(m,q)}^{(t)}$ approximates the contribution of sample $q$ within a task $m$, the task-level $p_{m}^{(t)}$ approximates the contribution of a task $m$. Accordingly, $p_{(m,q)}^{(t)}$ is the average absolute group-normalized advantage over a rollout group $G_{(m,q)}$ corresponding to a sample $q$ of a task $m$ at training step $t$. 
\begin{equation}
\small
p_{(m,q)}^{(t)}
\;=\;
\frac{1}{|G_{(m,q)}|}
\sum_{i\in G_{(m,q)}} \big|\hat A_{(m,q,i)}^{(t)}\big|
\end{equation}

% \chan{It would be good to intuitively explain why this is a good metric. Currently its ad hoc. (I understand that from our discussions something has come out, but we need to write it more intuitively)}

Then, we define $p_{m}^{(t)}$ as the average absolute group-normalized advantage over all rollouts collected for task $m$, with $\mathcal Q_m^{(t)}$ denoting the set of samples drawn for task $m$:
\begin{equation}
\small
p_{m}^{(t)}
\;=\;
\frac{
\sum_{q\in \mathcal Q_m^{(t)}} \sum_{i\in G_{(m,q)}} \big|\hat A_{(m,q,i)}^{(t)}\big|
}{
\sum_{q\in \mathcal Q_m^{(t)}} |G_{(m,q)}|
}
\end{equation}

\textbf{Structured modulation via a geometric reference.}
We leverage contribution signals to rebalance influence on the policy update, amplifying samples and tasks with lower contribution signals, and attenuating those with higher ones. Accordingly, we measure how far the contribution signal of each sample and task lay above or below a geometric-mean reference, and proportionately downscale or upscale their associated advantages. On the one hand, taking reference from the geometric-mean ensures that scaling is directly comparable across samples and tasks, and does not change the overall update size. On the other hand, it yields a ratio-based construction of scaling factors that tempers excessive variations in values (as we observe that contribution signals can vary by orders of magnitude, App.~\ref{app:additional_plots} Fig.~\ref{fig:appendix_add_factors}).
% naturally dampens extreme variations, preventing excessively large scaling factors.
% While the use of a geometric-mean is motivated by ensuring that scaling are directly comparable, the operation in the log space reduces excessively large variations in contribution signals. 
% Samples and tasks with lower contributions are amplified, while those with higher contributions are attenuated.
% To implement this rebalancing, the proxy signals $\{p_{(m,q)}^{(t)}, p_m^{(t)}\}$ are converted into sample and task-level modulation factors $\{s_{(m,q)}^{(t)}, s_m^{(t)}\}$, which scale group-normalized advantages.
% to balance influence on the policy update.
% \chan{It may make sense to bring up this intuition. Else I think the revs can get lost in the math.}
% To enable directly comparable and proportionate scaling across samples and tasks, without changing the overall update size, we measure how each contribution estimate deviates from a geometric-mean reference.
% We do so in the log space to mitigate excessively large variations in contribution magnitudes. 

% Concretely, we define a geometric mean reference at both the sample-level, $\bar p_{\mathrm{ref},m}^{(t)}$, and the task-level, $\bar p_{\mathrm{ref},\mathcal M}^{(t)}$:
% \begin{equation}\label{eq:geom_ref}
% \begin{aligned}
% \bar p_{\mathrm{ref},m}^{(t)}
% &=
% \Big(\prod_{q \in \mathcal Q_m^{(t)}} p_{(m,q)}^{(t)}\Big)^{\frac{1}{|\mathcal Q_m^{(t)}|}}. \\
% \bar p_{\mathrm{ref},\mathcal M}^{(t)}
% &=
% \Big(\prod_{m \in \mathcal M} p_m^{(t)}\Big)^{\frac{1}{|\mathcal M|}}
% \end{aligned}
% \end{equation}

Concretely, we define a geometric mean reference at both the sample-level, $\bar p_{\mathrm{ref},m}^{(t)}$, and the task-level, $\bar p_{\mathrm{ref},\mathcal M}^{(t)}$:

\begin{equation}\label{eq:geom_ref}
\small
\begin{aligned}
\bar p_{\mathrm{ref},m}^{(t)}
&=
\left(\prod_{q \in \mathcal Q_m^{(t)}} p_{(m,q)}^{(t)}\right)^{\!\frac{1}{|\mathcal Q_m^{(t)}|}}
\;
\bar p_{\mathrm{ref},\mathcal M}^{(t)}
=
\left(\prod_{m \in \mathcal M} p_m^{(t)}\right)^{\!\frac{1}{|\mathcal M|}}
\end{aligned}
\end{equation}

% This establishes common baselines for evaluating each sample’s contribution relative to other samples within the same task, and each task’s contribution relative to all tasks. the deviation of each sample and task contribution signal from these baselines. 

This establishes common baselines for comparing contributions across samples within a task or across tasks overall. For each sample or task contribution signal, we take the reciprocal of its ratio to the geometric-mean. This yields modulation factors $s_{(m,q)}^{(t)}$ at the sample-level, and $s_{m}^{(t)}$ at the task-level. 
% These deviations are constructed in the log space to temper excessively large relative differences, since we observe that $p^{(t)}$ can vary by orders of magnitude (see App.~\ref{}). Exponentiation then converts these deviations into positive modulation factors, ensuring that advantages are rescaled in proportion to their deviation from the reference, without changing their sign ($s_{(m,q)}^{(t)}$ at the sample-level; $s_{m}^{(t)}$ at the task-level):

\begin{equation}\label{eq:reciprocal_ratio}
\small
\begin{aligned}
s_{(m,q)}^{(t)}
&=
\frac{\bar p_{\mathrm{ref},m}^{(t)}}{p_{(m,q)}^{(t)}}
\qquad
s_m^{(t)}
=
\frac{\bar p_{\mathrm{ref},\mathcal M}^{(t)}}{p_m^{(t)}}
\end{aligned}
\end{equation}

The factors scale the group-normalized advantages of each rollout, yielding the \names\ advantage:
\begin{equation}
\small
A^{\mathrm{H}}_{(m,q,i)}{}^{(t)}
\triangleq
s_{(m,q)}^{(t)} \, s_{m}^{(t)} \, \hat A_{(m,q,i)}^{(t)}
\end{equation}
Due to their reciprocal construction, the modulation factors act to balance advantage magnitudes across samples and tasks. Specifically, within a task, samples whose contribution signals exceed the sample reference (i.e., $p_{(m,q)}^{(t)} > \bar p_{\mathrm{ref},m}^{(t)}$) receive modulation factors $s_{(m,q)}^{(t)} < 1$, while those below the reference (i.e., $p_{(m,q)}^{(t)} < \bar p_{\mathrm{ref},m}^{(t)}$) receive $s_{(m,q)}^{(t)} > 1$. Consequently, within the same task, stronger samples have downscaled advantages while weaker samples have upscaled advantages. Analogously, at the task-level, the advantages of tasks with contribution signals above the reference $\bar p_{\mathrm{ref},\mathcal M}^{(t)}$ are downscaled, while those below the reference are upscaled.
Additionally, as the modulation factors are constructed from a geometric mean reference, the factors have a geometric mean of 1.  
% Specifically, for each task m,
% \prod_{q \in \mathcal Q_m^{(t)}} s_{(m,q)}^{(t)}
% =
% \prod_{q \in \mathcal Q_m^{(t)}} \frac{\bar p_{\mathrm{ref},m}^{(t)}}{p_{(m,q)}^{(t)}}
% =
% \frac{(\bar p_{\mathrm{ref},m}^{(t)})^{|\mathcal Q_m^{(t)}|}}{\prod_{q \in \mathcal Q_m^{(t)}} p_{(m,q)}^{(t)}}
% =
% \frac{\prod_{q \in \mathcal Q_m^{(t)}} p_{(m,q)}^{(t)}}{\prod_{q \in \mathcal Q_m^{(t)}} p_{(m,q)}^{(t)}}
% =
% 1,
% where we used (\bar p_{\mathrm{ref},m}^{(t)})^{|\mathcal Q_m^{(t)}|}
% = \prod_{q \in \mathcal Q_m^{(t)}} p_{(m,q)}^{(t)} by definition of the geometric mean. Likewise, across tasks,
% \prod_{m \in \mathcal M} s_m^{(t)}
% =
% \prod_{m \in \mathcal M} \frac{\bar p_{\mathrm{ref},\mathcal M}^{(t)}}{p_m^{(t)}}
% =
% \frac{(\bar p_{\mathrm{ref},\mathcal M}^{(t)})^{|\mathcal M|}}{\prod_{m \in \mathcal M} p_m^{(t)}}
% =
% 1.
($\prod_{q \in \mathcal Q_m^{(t)}} s_{(m,q)}^{(t)} = 1$ 
and 
$\prod_{m \in \mathcal M} s_m^{(t)} = 1$, App.~\ref{app:harpo_full_formulation} shows the full derivation). This ensures that multiplicative upscaling from certain modulation factors are exactly compensated by downscaling from others. Thus, the factors cannot simultaneously enlarge or shrink all advantages at the sample or task-level, mitigating unintended influence on the global step size. 

% \chan{Intuition makes sense might want to bring it up}

% $p_c^{(t)} > \bar p_{\mathrm{ref}}^{(t)}$ implies $s_c^{(t)} < 1$, while $p_c^{(t)} < \bar p_{\mathrm{ref}}^{(t)}$ implies $s_c^{(t)} > 1$, thereby shrinking dominant components and amplifying weaker ones to promote more balanced optimization. Additionally, because of geometric centering in the log space, $\sum_{c\in\mathcal C}\Delta_c^{(t)} = 0$, which implies the resulting modulation factors are multiplicatively centered to 1, or equivalently, $\prod_{c\in\mathcal C} s_c^{(t)} = \exp\!\Big(\sum_{c\in\mathcal C}\Delta_c^{(t)}\Big) = 1.$ This ensures that the factors cannot simultaneously enlarge or shrink all components' advantages, mitigating unintended influence to the global step-size. 

\textbf{Inertial control for stable modulation.}
The modulation mechanism is updated on a slower time scale than the policy parameters. Accordingly, we maintain inertial estimates of the contribution signals and the modulation factors, allowing modulation to respond to persistent trends in contribution signals, rather than stochastic on-policy fluctuations. Contribution signals are smoothed using an exponential moving average to mitigate stochastic noise~\citep{kingma2014adam}. Modulation factors, as multiplicative ratios, are smoothed via multiplicative updates rather than by additive adjustments~\citep{arora2012multiplicative, bubeck2015convex}. For notational simplicity, we use $p^{(t)}$ and $s^{(t)}$ to denote both the newly computed quantities from the current rollout batch and their recursively updated running estimates, with the latter serving as the operational quantities used in modulation.

\begin{equation}\label{eq:inertial_control}
\small
\begin{aligned}
p_{(m,q)}^{(t)} 
&= \beta_\rho\,p_{(m,q)}^{(t-1)} + (1-\beta_\rho)\,p_{(m,q)}^{(t)} \\
p_{m}^{(t)} 
&= \beta_\rho\,p_{m}^{(t-1)} + (1-\beta_\rho)\,p_{m}^{(t)} \\
s_{(m,q)}^{(t)} 
&= \left(s_{(m,q)}^{(t-1)}\right)^{\beta_s}
   \left(s_{(m,q)}^{(t)}\right)^{1-\beta_s} \\
s_{m}^{(t)} 
&= \left(s_{m}^{(t-1)}\right)^{\beta_s}
   \left(s_{m}^{(t)}\right)^{1-\beta_s}
\end{aligned}
\end{equation}

% This inertial smoothing ensures that the modulation factors respond to persistent trends in contribution signals, rather than to stochastic on-policy fluctuations in their values.

% \chan{People may ask for abalations here.}

\textbf{\names\footnote{App.~\ref{app:algo} Algorithm~\ref{alg:harpo_full} summarizes HARPO. Additional details of reward design are in App.~\ref{app:reward_full_formulation}} objective.} With the HARPO-modulated advantage in place, \names\ retains the PPO-style trust-region objective of GRPO, substituting $A^{\mathrm{H}}$ for $\hat A$ to construct the clipped surrogate $\tilde A^{\mathrm{H}}_{(m,q,i):k}(\theta)$, thereby forming the HARPO objective:
\begin{equation}
\small
\label{eq:harpo_compact_controlled}
\begin{aligned}
J_{\mathrm{\names}}(\theta)
&=
\mathbb{E}_{(m,q)\sim \mathcal D}
\mathbb{E}_{\{o_{(m,q,i)}\}\sim \pi_{\theta_{\mathrm{old}}}}\!\Bigg[
\frac{1}{|G_{(m,q)}|}
\sum_{i\in G_{(m,q)}}
\\
&\hspace{-60pt}\phantom{=\mathbb{E}\Bigg[}
\frac{1}{n_{o_{(m,q,i)}}}
\sum_{k=1}^{n_{o_{(m,q,i)}}}
\tilde A^{\mathrm{H}}_{(m,q,i):k}(\theta)
\Bigg]
\;-\;
\beta\,
\mathbb{E}\!\left[
D_{\mathrm{KL}}\!\left(\pi_\theta\;\|\;\pi_{\mathrm{ref}}\right)
\right]
\end{aligned}
\end{equation}

% \paul{where does signal v noise come from? not motivated}

% \paul{do we really need all these 3 components? also can u come up with a unified approach that captures all 3 of them? its just different weighting strategies right - right now it reads very hacky}

% \keane{Algorithm Block}

\subsection{Reward Design}
\label{sec:reward_design}

% We use a unified reward design\footnote{Full details of reward design in App.~\ref{}, including reward formulas, response formats, parameters}. that includes rewards for accuracy, format compliance, and response length.

\textbf{Accuracy.} For classification tasks, we use a binary reward $r_{\mathrm{cls}}$, which equals 1 for an exact label match and 0 otherwise. For question-answering tasks with free-text responses, we use cosine similarity rewards, $r_{qa}$, normalized to a scale of $[0,1]$ for compatibility with rewards assignment, to measure alignment between the generated response and answer.
% The responses and answers are embedded by a lightweight language model, and the cosine rewards are normalized to a scale of $[0,1]$, to ensure compatibility with the rewards structure in policy optimization.

 % (computed using MiniLM)
\textbf{Formatting.}
We add a binary reward, $r_{\mathrm{fmt}}$, for adhering to the response structure, which includes reasoning traces followed by the model's prediction.

\textbf{Length.}
Following~\citet{zhang2025grpo_lead}, we leverage an overlong length penalty, $r_{\mathrm{len}}$, to prevent excessive verbosity of responses.

\textbf{Final reward}\footnotemark[\value{footnote}]\textbf{.}
With format weight $w_{\mathrm{fmt}}=0.2$ and length scale $\lambda_{\mathrm{len}}=0.75$, the final per-sample reward is the following, where $r_{task}$ can be $r_{cls}$ or $r_{qa}$, depending on if the sample belongs to a classification or QA task respectively. 
\[
r=(1-w_{\mathrm{fmt}})\,r_{\mathrm{task}} + w_{\mathrm{fmt}}\,r_{\mathrm{fmt}} + \lambda_{\mathrm{len}}\,r_{\mathrm{len}},\]

\begin{table*}[h]
\small
\centering
\vspace{-10pt}\caption{
% We provide the full definitions of each behavioral task (EMO, HUM, INT etc.) in App.~\ref{}
Per-task performance (\%) is reported for both models and training algorithms. Each task may include multiple datasets; we report the mean performance across a task's associated datasets, with full results in App. Tab.~\ref{tab:full_dataset_results}. Following~\citet{ong2025human}, we report binary weighted F1 for SEN; mean per-class weighted accuracy for EMO; weighted F1 for HUM, SAR, ANX, DEP, and PTSD; and LLM-Judge accuracy for SOC, INT, and NVC. Given the disparities between the metrics, we compute average performance rank across tasks separately for models and algorithms ($\downarrow$ indicates lower score is better), with ties assigned the same rank. Best results are bolded and second-best are underlined. (*) denotes existing models that have been trained on more than two behavioral tasks, reflecting substantive multitask training; only a few such models are publicly available. For fair comparison, all training algorithm results reflect training on the same Human Behavior Atlas benchmark, using an identical reward design (Sec.~\ref{sec:reward_design}) and the same base model (Qwen2.5-Omni-7B).}
\setlength{\tabcolsep}{7pt}
\renewcommand{\arraystretch}{0.95}
\resizebox{\linewidth}{!}{%
\begin{tabular}{lcccccccccc|c}
\toprule
\textbf{Models}
& \textbf{EMO}
& \textbf{HUM}
& \textbf{INT}
& \textbf{PTSD}
& \textbf{ANX}
& \textbf{DEP}
& \textbf{SEN}
& \textbf{SAR}
& \textbf{SOC}
& \textbf{NVC}
& \textbf{Avg. Rank} $\downarrow$ \\
\midrule
Gemma 4-E4B (8B)
& 55.98 & 52.63 & 40.25 & 86.00 & 33.36 & 22.55 & 69.55 & \underline{73.54} & 20.40 & 4.65 & 7.30 \\
Gemma-3-4B
& 55.03 & 59.70 & 22.70 & 49.90 & 60.10 & 46.25 & 73.83 & 52.90 & 19.10 & 2.30 & 8.30 \\
Qwen~2.5-Omni-7B
& 58.25 & 54.30 & 25.40 & 76.00 & 79.30 & 71.35 & 67.20 & 65.60 & 25.40 & 6.90 & 6.20 \\
Qwen~2.5-VL-7B
& 54.08 & 58.30 & 24.90 & 75.50 & 63.10 & 63.80 & 50.50 & 51.10 & 23.10 & 9.80 & 7.80 \\
Qwen~3-VL-8B-Instruct
& 57.66 & \underline{66.76} & 38.00 & 92.70 & 42.29 & 51.62 & 69.70 & 63.67 & 24.94 & 13.95 & 5.50 \\
HumanOmniV2-7B$^{*}$
& 59.70 & 63.80 & 26.30 & 82.40 & 52.70 & 65.40 & 74.20 & 39.50 & \underline{28.20} & 9.30 & 5.80 \\
OmniSapiens SFT$^{*}$
& 63.08 & 53.20 & 25.60 & \textbf{100.00} & 90.90 & 73.25 & 76.77 & 62.40 & 25.70 & 12.10 & 4.40 \\
OmniSapiens BAM$^{*}$
& \underline{64.53} & 64.40 & 17.70 & \textbf{100.00} & 90.90 & \underline{78.85} & \textbf{78.53} & \textbf{79.50} & 20.10 & \textbf{16.20} & \underline{3.30} \\
OmniSapiens-7B RL$^{*}$
& 57.28 & 63.90 & \underline{48.60} & 96.80 & \underline{91.90} & 77.15 & 39.60 & 64.70 & \textbf{30.40} & 13.30 & 4.20 \\
\midrule
\textbf{OmniSapiens-7B 2.0 (ours)} 
& \textbf{76.55} 
& \textbf{69.85} 
& \textbf{50.52} 
& \underline{98.39} 
& \textbf{91.98} 
& \textbf{78.87} 
& \underline{77.61} 
& 70.64 
& 25.40 
& \underline{14.54} 
& \textbf{1.90} \\
\midrule\midrule
\multicolumn{12}{l}{\textbf{Training Algorithms}} \\
\midrule
RLOO~\citep{ahmadian2024rloo}
& 75.58
& 67.86
& 51.73
& \textbf{98.39}
& 90.68
& 77.57
& 76.86
& 62.58
& \underline{29.54}
& \textbf{16.28}
& \underline{2.80} \\
RE++~\citep{hu2025reinforce++}
& 75.92
& 60.26
& 5.01
& \textbf{98.39}
& \textbf{93.11}
& 73.87
& 56.52
& 50.21
& 12.64
& 4.07
& 4.50 \\
GPG~\citep{chu2025gpg}
& \textbf{77.70}
& \underline{69.28}
& \textbf{54.21}
& \textbf{98.39}
& 90.40
& \underline{78.40}
& 75.77
& 45.96
& 27.93
& 12.79
& 2.90 \\
GRPO~\citep{shao2024deepseekmath}
& 76.45
& 27.56
& 49.90
& \textbf{98.39}
& 90.40
& 77.64
& \underline{77.51}
& 53.58
& 23.30
& 11.00
& 3.90 \\
EMAGRPO
& 73.36
& 63.50
& \underline{52.62}
& \textbf{98.39}
& 90.68
& 73.36
& 68.28
& \textbf{77.75}
& \textbf{30.58}
& \underline{15.70}
& 3.10 \\
\midrule
\textbf{\names\ (ours)} 
& \underline{76.55} 
& \textbf{69.85} 
& 50.52 
& \textbf{98.39} 
& \underline{91.98} 
& \textbf{78.87} 
& \textbf{77.61} 
& \underline{70.64} 
& 25.40 
& 14.54 
& \textbf{2.10} \\
\bottomrule
\end{tabular}
}
\label{tab:models_and_algos_multi_task_results}
\vspace{-9.5pt}
\end{table*}

% \textbf HARPOv2
% & \underline{77.59}
% & 66.37
% & \underline{52.11}
% & \textbf{98.39}
% & \textbf{93.52}
% & \underline{77.42}
% & \textbf{79.17}
% & \textbf{63.21}
% & \textbf{31.61}
% & \underline{13.95}
% & \textbf{1.70} \\

% \textbf{HARPOv4 (ours)}
% & \underline{77.33}
% & \underline{68.84}
% & \underline{51.97}
% & \textbf{98.39}
% & \underline{92.58}
% & 77.42
% & 76.62
% & \textbf{81.64}
% & \textbf{30.58}
% & \textbf{16.86}
% & \textbf{1.90} \\

% \textbf{\names\ (ours, no EMA)}
% & \underline{77.47}
% & \textbf{71.50}
% & \underline{52.86}
% & \textbf{98.39}
% & 91.89
% & 76.67
% & 70.79
% & \underline{74.89}
% & \underline{29.89}
% & \underline{15.70}
% & - \\
% \textbf{\names\ (ours, unnormalized)}
% & 76.32
% & 63.46
% & 51.64
% & \textbf{98.39}
% & \textbf{94.36}
% & \textbf{79.09}
% & 77.11
% & \textbf{83.41}
% & \textbf{30.81}
% & 15.70
% & - \\

\section{Experiments}
% \paul{dont jump into the benchmark and models yet, state the problem and define notation first}

Leveraging the Human Behavior Atlas benchmark~\citep{ong2025human}, we analyze the capabilities of \namesm\ as a foundation model for social behavior processing. We conduct our evaluation across three key dimensions. First, we assess \textit{multitask capability}, evaluating whether a single model can coherently execute diverse behavioral analysis tasks, thereby obviating the need to train, store, and deploy separate task-specific models. Second, we evaluate \textit{zero-shot generalization}, assessing whether learned representations transfer to unseen behavioral contexts without task-specific supervision. Such transfer mitigates annotation costs that typically constrain deployment to novel domains. Third, we analyze the \textit{reasoning trace interpretability}, which crucially enables post-hoc verification of model judgements in critical behavioral applications (i.e., health).

% Third, we analyze reasoning trace quality as an indicator of how interpretable the model's decisions are, which is especially required in critical behavioral analysis applications (i.e., mental health)

% We use the Human Behavior Atlas benchmark~\citep{ong2025human} for training and evaluation, which covers diverse human behavior tasks with over 100k samples. The tasks include sentiment polarity (SEN), emotion recognition (EMO), social reasoning (SOC), intent recognition (INT), non-verbal communication (NVC), as well as detecting humor (HUM), sarcasm (SAR), anxiety (ANX), depression (DEP), and PTSD (PTSD). Given the multimodal nature of the data (text, vision, and audio), we adopt Qwen 2.5 Omni-7B~\citep{xu2025qwen2.5omni} as our base architecture for \namesm. All training follows a multitask setup, where a single model is jointly trained across all tasks.

% as our base model for \namesm
% We additionally train a vision-only variant of \namesm based on the Qwen3-VL-8B-Instruct architecture, denoted as \names-VL; its results are reported in App.~\ref{app:vl_results}.

\subsection{Multitask Performance}

RQ1: We evaluate \namesm\ against: (1) general-purpose foundation model families: Gemma~\cite{team2025gemma3}, Qwen~\cite{xu2025qwen3}; and (2) specialized models trained on human behavior data, comprising HumanOmniV2~\cite{yang2025humanomniv2} and prior Omnisapiens variants (SFT, BAM, RL)~\cite{ong2025human}. Using the same model backbone, Qwen 2.5-Omni~\cite{xu2025qwen2.5omni}, we also compare \names\ against current critic-free RL methods that explicitly address learning heterogeneous data, (1) normalization-based: GRPO~\citep{shao2024deepseekmath}, RE++~\citep{hu2025reinforce++}, (2) variance-reduction: RLOO~\citep{ahmadian2024rloo}, (3) unbiased gradients: GPG~\citep{chu2025gpg}, (4) multitask balancing: EMAGRPO~\citep{feng2025emagrpo}.

\textbf{Comparing model performance}. \namesm\ achieves state-of-the-art performance for multitask social behavior processing. Relative to existing models, \namesm\ achieves the strongest overall performance across multiple behavioral tasks. From Tab.~\ref{tab:models_and_algos_multi_task_results}, \namesm\ outperforms general-purpose LLMs (Gemma-3, Gemma-4, Qwen-2.5-Omni, Qwen-2.5-VL, Qwen-3-VL). It also outperforms prior behavioral foundation models for diverse social-behavioral tasks spanning multiple behavioral dimensions, including HumanOmniV2-7B and OmniSapiens-7B RL, BAM, SFT. In particular, \namesm\ ranks in the top 2 on 8 of 10 tasks, with the strongest average task performance rank (1.90).

\textbf{RL algorithm performance comparison}. Among state-of-the-art reasoning RL algorithms, \names\ achieves the most consistent performance across behavioral tasks. As shown in Tab.~\ref{tab:models_and_algos_multi_task_results}, \names\ outperforms reasoning-based RL methods spanning different approaches, normalization: GRPO~\cite{shao2024deepseekmath}, RLOO~\cite{ahmadian2024rloo}, variance reduction: RE++~\cite{hu2025reinforce++}, unbiased gradients: GPG~\cite{chu2025gpg}, and multitask balancing: EMA-GRPO~\cite{feng2025emagrpo}, attaining the best average rank (2.10) across tasks. While GRPO, GPG, and RE++ performance collapse on certain tasks (e.g., SAR), \names\ does not exhbit the same performance degradation on weaker tasks, indicating more balanced multitask learning.

\subsection{Zero-Shot Generalization Performance}~\label{sec:generalization}
\vspace{-20pt}

% \begin{table}[h]
% \small
% \centering
% \caption{
% Zero-shot generalization weighted F1 performance (\%) on AUT and SER, compared to existing models.
% }
% \setlength{\tabcolsep}{10pt}
% \renewcommand{\arraystretch}{0.95}
% \resizebox{0.9\linewidth}{!}{%
% \begin{tabular}{lcc}
% \toprule
% \textbf{Model}
% & \textbf{AUT}
% & \textbf{SER} \\
% \midrule
% Qwen 2.5-Omni-7B
% & 25.68 & 53.53 \\
% OmniSapiens-7B RL
% & 30.46 & 55.77 \\
% HumanOmniV2-7B
% & \underline{38.05} & \underline{62.74} \\
% \midrule
% \textbf{OmniSapiens-7B 2.0 (ours)}
% & \textbf{39.91} & \textbf{72.11} \\
% \bottomrule
% \end{tabular}
% }
% \label{tab:aut_esr_results}
% \end{table}

RQ2: We evaluate zero-shot generalization of \namesm\ on five diverse held-out behavioral tasks, without further fine-tuning: (1) Autism Detection (AUT)~\cite{deng2024avasd}; (2) Speech Emotion Recognition (SER)~\cite{Busso2008IEMOCAP:Interactiveemotionaldyadic}; (3) In-the-wild depression recognition (IDR)~\cite{shen2022automaticeatd}; (4) Social Media Sentiment Analysis (SMSA)~\cite{niu2016sentimentmvsd}; (5) Sarcasm Incongruity Recognition (SIR)~\cite{yue2024sarcnet}.

\begin{table}[h]
\small
\centering
\vspace{-10pt}
\caption{
Zero-shot generalization weighted F1 performance (\%) on heldout tasks AUT, SER, IDR, SMSA, and SIR. *OmniSapiens-7B SFT and BAM are excluded, as their task-specific classifier heads structurally preclude zero-shot inference.
}
\setlength{\tabcolsep}{6pt}
\renewcommand{\arraystretch}{0.95}
\resizebox{\linewidth}{!}{%
\begin{tabular}{lccccc}
\toprule
\textbf{Model}
& \textbf{AUT}
& \textbf{SER}
& \textbf{IDR}
& \textbf{SMSA}
& \textbf{SIR} \\
\midrule
Qwen 2.5-Omni-7B
& 25.68 & 53.53 & \underline{70.25} & 44.64 & 34.99 \\
Gemma 4-E4B-Instruct (8B)
& 16.78 & 58.30 & 11.37 & 55.60 & 56.74 \\
OmniSapiens-7B RL
& 30.46 & 55.77 & 69.29 & \underline{55.03} & \underline{66.53} \\
HumanOmniV2-7B
& \underline{38.05} & \underline{62.74} & 21.97 & 53.06 & 37.45 \\
\midrule
\textbf{OmniSapiens-7B 2.0 (ours)}
& \textbf{39.91} & \textbf{72.11} & \textbf{72.43} & \textbf{58.47} & \textbf{69.27} \\
\bottomrule
\end{tabular}
}
\label{tab:aut_esr_results}
\vspace{-15pt}
\end{table}

\textbf{Zero-shot generalization performance.}
From Tab.~\ref{tab:aut_esr_results}, \namesm\ achieves the best performance across all five tasks spanning a broad spectrum of behavioral contexts, including clinical behavior (AUT, IDR), affective expression (SER), and social-pragmatic reasoning (SMSA, SIR). The uniform gains across held-out tasks indicate that relative to existing models, \namesm\ learns behavioral representations that can more effectively generalize and transfer beyond training datasets and distributions. As a result, \namesm\ demonstrates stronger effectiveness than existing models when applied to new datasets and distributions, as frequently encountered in real-world deployment settings.

\textbf{Comparison with OmniSapiens-7B RL}. Despite being trained on the same Human Behavioral Atlas benchmark, \namesm, trained with \names\ instead of the standard GRPO, exhibits stronger generalization performance than OmniSapiens-7B RL across each held-out task. This coincides with improved multitask performance achieved by \namesm\ from Tab.~\ref{tab:models_and_algos_multi_task_results}, with an average task rank of 1.90 compared to 4.20 for OmniSapiens-7B RL. This suggests that more uniform learning across behavioral tasks through \names\ may promote the acquisition of transferable behavioral features, yielding improved generalization. Though we leave a more rigorous investigation of this relationship to future work.

% transferable
% This suggests that the \names-trained \namesm\ avoids concentrating learning on a specific tasks, instead promoting more uniform learning across behavioral tasks. Such balanced optimization appears to support the acquisition of more transferable behavioral features, contributing to improved generalization performance.

% SER evaluates emotion recognition from speech under different labeling schemes and conversational settings than those seen during training. 

% We evaluate generalization on AV-ASD and IEMOCAP to assess robustness across distinct behavioral domains and data conditions. AV-ASD represents a data-scarce behavioral setting, where high annotation costs and privacy constraints limit the availability of large-scale labeled data. This makes it a suitable testbed for examining whether a general-purpose behavioral model can perform effectively without relying on extensive in-domain supervision.

% IEMOCAP, in contrast, focuses on emotion recognition from speech. Although our model has been trained on emotion-related datasets with some distributional overlap, we use IEMOCAP to evaluate robustness under different annotation protocols, contextual settings, and interaction styles. Prior work has shown that emotion and affective models often exhibit limited generalization across such shifts, making this an informative benchmark for assessing transfer performance.

% \midrule
% Qwen 2.5-Omni-7B
% & & \\
% Qwen 3 VL-8B
% & & \\

% \vspace{-20pt}
\subsection{Reasoning Analysis}

% Behavioral analysis models have critical applications such as behavioral health, content moderation, amongst others, where interpretability is essential for understanding and validating model decisions.

RQ3: We analyze the interpretability of \namesm's reasoning traces\footnote{We make public the reasoning traces, and provide the experimental setup for the consistency and paraphrase analysis, and human evaluation instructions, annotators in App.~\ref{app:reasoning_eval_setup}.}, using the held-out datasets in Sec.~\ref{sec:generalization} to evaluate under realistic deployment conditions, and to control for training data overlap. We evaluate interpretability through a multi-dimensional notion of explanation quality, grounded in established XAI desiderata~\citep{langer2021xai,nauta2023anecdotalxai,turbe2023evaluation,wei2024revisiting,ong2025explainable}. We capture properties \textit{robustness}, \textit{efficiency} via automated metrics, consistency and mean token length); properties \textit{understandability}, \textit{completeness}, \textit{compactness} via human evaluation of coherence, specificity, and concision.

% Preamble (if not already):
% \usepackage{graphicx}

% RQ3: Beyond task performance, we compare reasoning traces\footnote{To support reproducibility and further study, we make publicly available all the reasoning traces after the review process.} for an interpretable view of the decision-making process of different methods and models. This provides insight into the robustness of the underlying inference behavior.

% \paul{include a bunch of examples of reasoning on hard social behaviors}

\begin{table}[h]
\small
\centering
\vspace{-10pt}
\caption{Reasoning evaluation results for robustness and efficiency of reasoning traces. Consistency is the fraction of inputs on which all 5 independently sampled reasoning traces produce the same answer. Average tokens is taken across the 5 sampled traces per input. Paraphrase consistency is the fraction of inputs on which the predicted answers agree across two semantically equivalent prompt variants. All metrics are averaged across all samples. ($\uparrow$ higher score is better; $\downarrow$ lower score is better)}
\setlength{\tabcolsep}{8pt}
\renewcommand{\arraystretch}{0.95}
\resizebox{\linewidth}{!}{%
\begin{tabular}{lrrr}
\toprule
Model 
& Consistency $\uparrow$ & Paraphrase $\uparrow$
& Avg. Tokens $\downarrow$ \\
\midrule
Qwen2.5-Omni-7B 
& 34.0 & 58.2 & 73.66 \\
Gemma 4-E4B (8B)
& \underline{66.5} & 79.2 & 211.10 \\
HumanOmniV2 
& 50.0 & 78.2 & 195.90 \\
OmniSapiens-7B-RL 
& 55.1 & \underline{80.7} & \underline{57.69} \\
\midrule
OmniSapiens 2.0 (ours) 
& \textbf{87.7} & \textbf{88.4} & \textbf{19.86} \\
\bottomrule
\end{tabular}%
}
\vspace{-6pt}
\label{tab:reasoning_eval_machine}
\end{table}

\textbf{Robustness and efficiency of reasoning.} Following prior work~\citep{wang2022self, jie2024interpretable}, robustness is assessed via two consistency measures: agreement across independently sampled reasoning traces (Consistency) and agreement across semantically equivalent prompt variants (Paraphrase). From Tab.~\ref{tab:reasoning_eval_machine}, \namesm\ achieves the strongest reasoning consistency (87.7 vs.\ 66.5 next best) and the highest paraphrase consistency (88.4 vs.\ 80.7 next best), indicating that its reasoning traces are both stable under stochastic sampling and robust to input variation. This consistency yields a more analyzable reasoning process, strengthening interpretability~\citep{alvarez2018towardsinterp}. Efficiency, measured by reasoning trace length (Avg. Tokens), reflects the cost of eliciting interpretability. \namesm\ produces more concise traces (19.86 tokens vs.\ 57.69 next best), reducing inference cost and the amount of information the reader must process.

% \begin{table}[h]
% \small
% \centering
% \caption{Human evaluation (\%). Raw Win: fraction preferred over baseline; Tie: equal preference; Win$^\dagger$: head-to-head win rate excluding ties.}
% \setlength{\tabcolsep}{1.0pt}
% \renewcommand{\arraystretch}{0.95}
% \resizebox{\linewidth}{!}{%
% \begin{tabular}{lccc|ccc|ccc}
% \toprule
% \textbf{\namesm\ vs. Baselines} 
% & \multicolumn{3}{c}{\textbf{Spec.}} 
% & \multicolumn{3}{c}{\textbf{Coh.}} 
% & \multicolumn{3}{c}{\textbf{Conc.}} \\
% \cmidrule(lr){2-4}\cmidrule(lr){5-7}\cmidrule(lr){8-10}
% & Win & Tie & Win$^\dagger$ 
% & Win & Tie & Win$^\dagger$ 
% & Win & Tie & Win$^\dagger$ \\
% \midrule
% Qwen2.5-Omni-7B 
% & 45.3 & 35.3 & 66.0 
% & 66.0 & 24.0 & 85.3 
% & 86.0 & 13.3 & 99.3 \\
% Gemma 4-E4B-Instruct (8B)
% & 59.3 & 16.7 & 70.0 
% & 76.0 & 12.0 & 85.4 
% & 98.0 & 0.7 & 98.7 \\
% HumanOmniV2 
% & 62.7 & 14.0 & 70.4 
% & 78.0 & 9.3 & 85.7 
% & 100.0 & 0.0 & 100.0 \\
% OmniSapiens-7B-RL 
% & 41.9 & 40.6 & 66.1 
% & 67.0 & 20.9 & 83.2 
% & 90.6 & 8.0 & 98.6 \\
% \midrule
% \textbf{Avg. Win Rates} 
% & 52.3 & 26.6 & 68.5 
% & 71.8 & 16.5 & 85.1 
% & 93.7 & 5.5 & 99.2 \\
% \bottomrule
% \end{tabular}%
% }
% \label{tab:human_eval}
% \end{table}

\begin{table}[t]
\small
\centering
\vspace{-5pt}
\caption{Human evaluation of \namesm\ reasoning trace interpretability over baselines. Evaluation criteria is based on specificity, coherence and concision of reasoning traces. Tie+Win: \% tied or preferred over baseline; Tie: \% equal preference; Win$^\dagger$: \% Win rate excluding ties. \namesm\ preferred over, or at the minimum tied with, every baseline across all three criteria.}
\setlength{\tabcolsep}{0.75pt}
\renewcommand{\arraystretch}{0.95}
\resizebox{\linewidth}{!}{%
\begin{tabular}{lccc|ccc|ccc}
\toprule
\textbf{vs. Baselines}
& \multicolumn{3}{c}{\textbf{Specificity}}
& \multicolumn{3}{c}{\textbf{Coherence}}
& \multicolumn{3}{c}{\textbf{Concision}} \\
\cmidrule(lr){2-4}\cmidrule(lr){5-7}\cmidrule(lr){8-10}
& Tie+Win & Tie & Win$^\dagger$
& Tie+Win & Tie & Win$^\dagger$
& Tie+Win & Tie & Win$^\dagger$ \\
\midrule
Qwen2.5-Omni-7B
& 80.6 & 35.3 & 66.0
& 90.0 & 24.0 & 85.3
& 99.3 & 13.3 & 99.3 \\
Gemma 4-E4B (8B)
& 76.0 & 16.7 & 70.0
& 88.0 & 12.0 & 85.4
& 98.7 & 0.7 & 98.7 \\
HumanOmniV2
& 76.7 & 14.0 & 70.4
& 87.3 & 9.3 & 85.7
& 100.0 & 0.0 & 100.0 \\
OmniSapiens-7B-RL
& 82.5 & 40.6 & 66.1
& 87.9 & 20.9 & 83.2
& 98.6 & 8.0 & 98.6 \\
\midrule
\textbf{Avg. Rates}
& 78.9 & 26.6 & 68.5
& 88.3 & 16.5 & 85.1
& 99.2 & 5.5 & 99.2 \\
\bottomrule
\end{tabular}%
}
\label{tab:human_eval}
\end{table}

\begin{figure}[h]
  \centering
\includegraphics[width=\columnwidth]{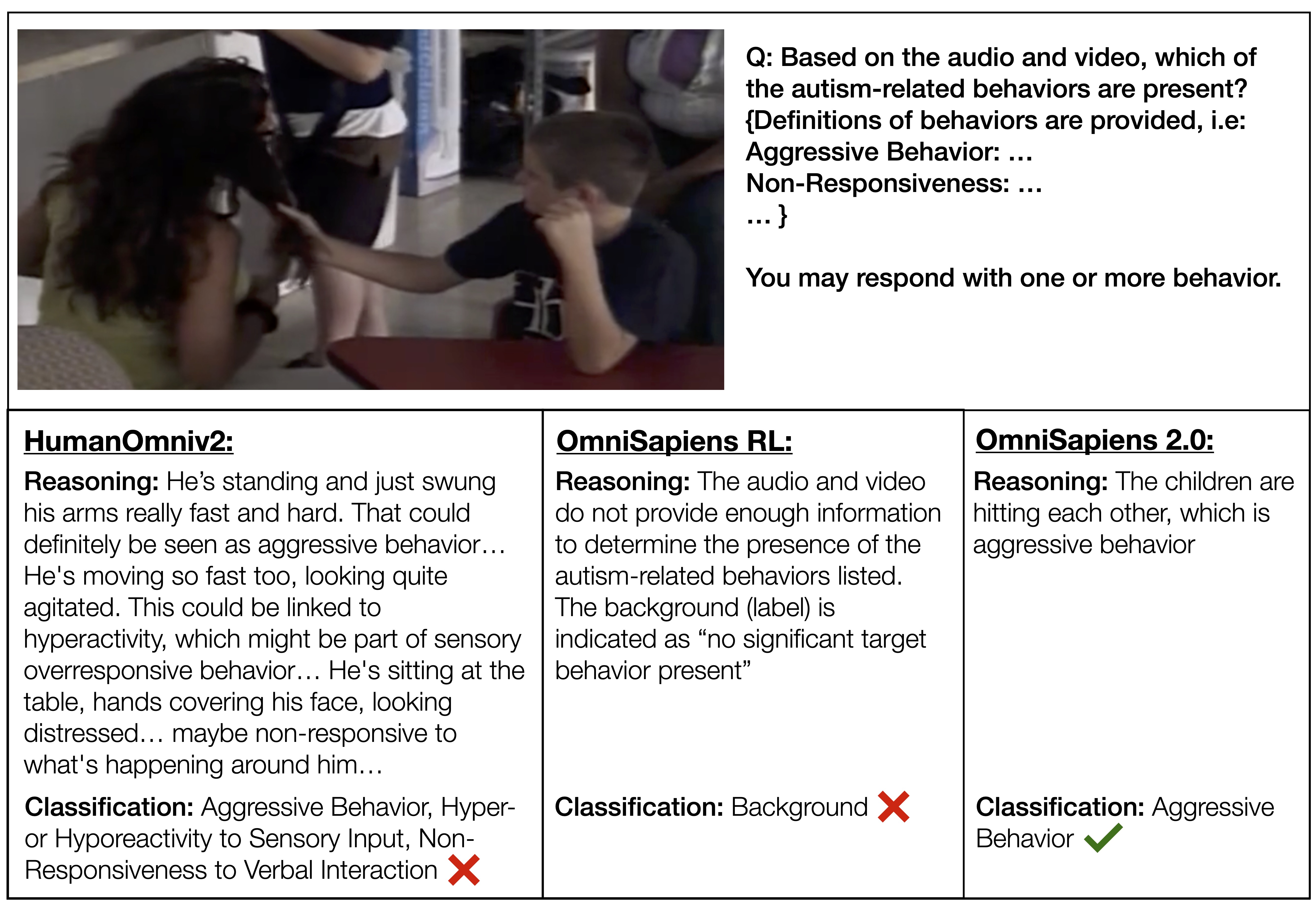}
\vspace{-18pt}
  \caption{Reasoning case study for autism behavioral detection (AUT); HumanOmniv2 produces verbose reasoning, OmniSapiens RL abstains from predictions, \namesm\ generates concise reasoning traces identifying the most salient behavior.}
  \label{fig:reasoning_case_study_avasd}
  \vspace{-15pt}
\end{figure}

\textbf{Human evaluation and case study of the completeness, understandability and compactness of reasoning.} To assess reasoning quality, annotators performed pairwise comparisons between \namesm\ and each baseline along three XAI desiderata: specificity (completeness), coherence (understandability), and concision (compactness). Averaged across baselines (Tab.~\ref{tab:human_eval}), \namesm\ is tied with or preferred over baselines in 78.9\%, 88.3\%, and 99.2\% of comparisons for specificity, coherence, and concision respectively; excluding ties yields effective win rates of 68.5\%, 85.1\%, and 99.2\%, indicating consistent preference for \namesm's more informative, logically structured, and succinct reasoning. A case study on the AUT task (Fig.~\ref{fig:reasoning_case_study_avasd}) against existing behavioral foundation models further illustrates these trends: HumanOmniV2 generates expansive chains that introduce behaviors without clear evidential support, while OmniSapiens-7B-RL exhibits the opposite failure mode, defaulting to ``background (no behavior)'' predictions even amid salient cues. \namesm, by contrast, produces concise traces focused on cues directly supported by the clip, neither over-predicting nor over-abstaining, consistent with the specificity, coherence, and concision preferences observed in the human evaluation.

\subsection{Empirical Analysis \& Ablations}\label{sec:ablations}
RQ4: To further evaluate HARPO, we conduct additional studies to examine its underlying mechanisms.

% \paul{check if any ablations recover baselines eg multitask RL, }

% --- HARPO ablations: per-task (template, no results yet) ---
% Requirements: \usepackage{booktabs}
% Optional: \usepackage{pifont} for \cmark/\xmark (not used here since results are blank)
% This is a two-column spanning table via table* and resizebox.

% which serves as a diagnostic of whether a task’s learning signal is amplified or suppressed relative to others during training

\begin{figure}[h]
    \centering
    \includegraphics[width=\linewidth]{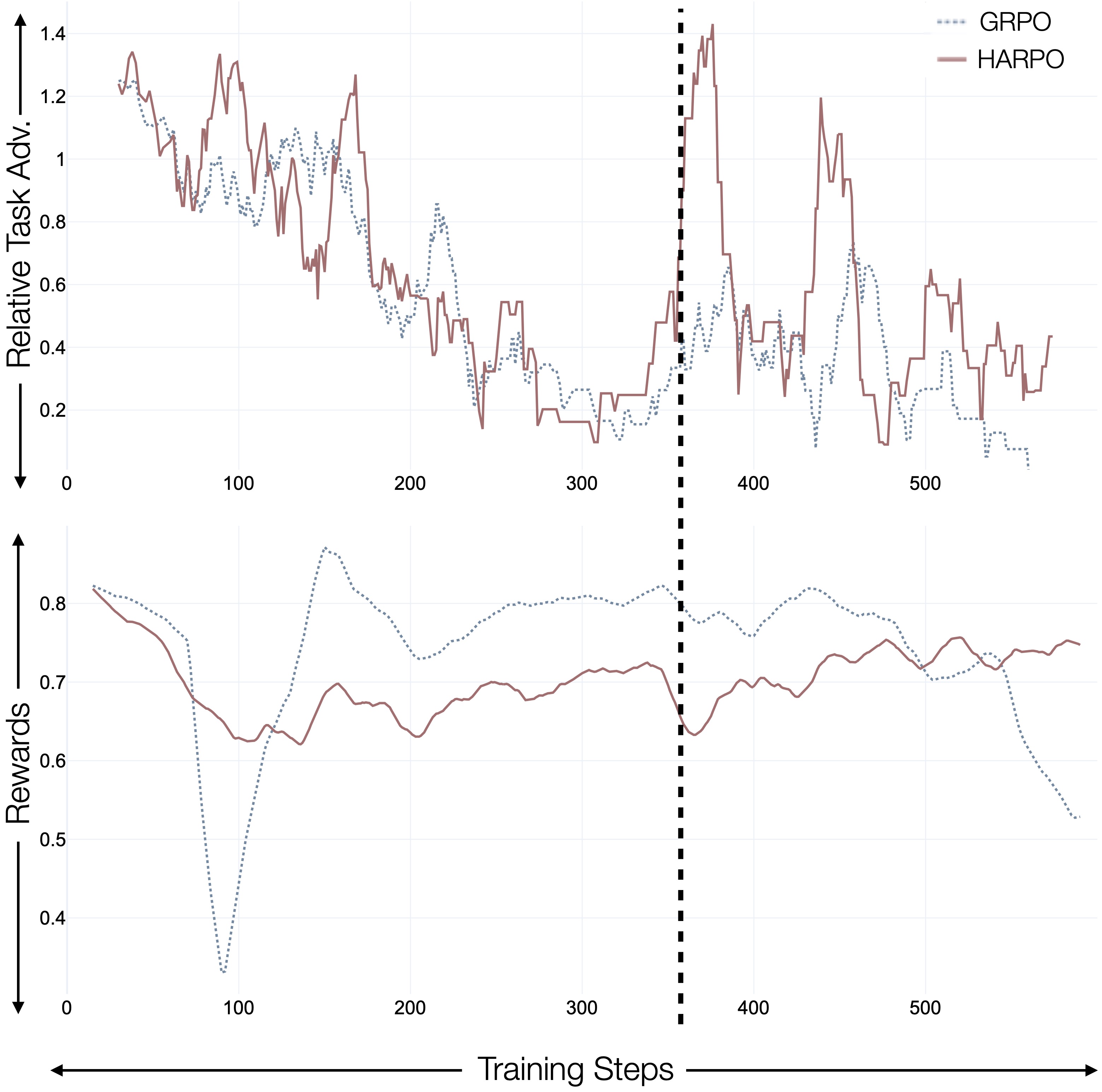}
    \vspace{-15pt}\caption{Comparison of relative task advantage and rewards for SAR. After step 355 (dotted line), HARPO exhibits increasing relative task advantage alongside increasing rewards, while GRPO maintains lower relative task advantage with decreasing rewards.}
    \label{fig:allocation_grpo_harpo}
    \vspace{-10pt}
\end{figure}

% \textbf{HARPOv2}
% & \underline{77.59}
% & 66.37
% & \underline{52.11}
% & \textbf{98.39}
% & \textbf{93.52}
% & \textbf{77.42}
% & \textbf{79.17}
% & 63.21
% & \textbf{31.61}
% & \underline{13.95}
% & \textbf{1.90} \\

% A central design principle of HARPO is to balance optimization by amplifying the advantage of tasks that contribute weakly to the shared policy update.

\textbf{HARPO’s advantage modulation on weak performing tasks.}
HARPO balances optimization by modulating advantages prior to the policy update step. To investigate this, we compute relative task advantage, defined as the ratio between a task’s average advantage magnitude and the mean of the average advantage magnitudes across all tasks, to observe how a change in this statistic affects learning.  We focus on SAR as a representative weak-performing task (performance of 53.58\% under GRPO; 70.64\% under HARPO), to more clearly study advantage modulation in this regime. From Fig.~\ref{fig:allocation_grpo_harpo}, in the later parts of training (i.e., after the dotted line depicting step 355), HARPO’s amplification of SAR’s relative task advantage coincides with a continued increase in average rewards. At the same training stage, GRPO's unmodulated relative task advantage remains lower, corresponding with a steady decline in rewards. This suggests that HARPO's advantage modulation sustains longer learning for weak-performing tasks with otherwise diminished advantages, contributing to improved performance.

 % We track how its relative task advantage evolves over 600 training steps, alongside task rewards

% \paragraph{Normalizing advantages by count prevents the proxy from being overwhelmed by sample proportions.}

% Structured modulation through log-space deviations and geometric centering prevents unintended scaling.

\begin{table}[t]
\small
\centering
\caption{
Average task performance rank for \names\ ablations  (lower is better), computed by ranking methods per task based on performance, then averaging ranks across tasks. We provide the per-task and per-dataset performance breakdown in App.~\ref{app:full_ablation_results}.
}
\setlength{\tabcolsep}{10pt}
\renewcommand{\arraystretch}{0.95}
\begin{tabular}{lc}
\toprule
\textbf{Variant} & \textbf{Avg. Rank} $\downarrow$ \\
\midrule
\textbf{\names} & \textbf{1.90} \\
w/o structured modulation & \underline{2.00} \\
w/o sample-level modulation & 2.60 \\
w/o inertial control & 2.70 \\
\bottomrule
\end{tabular}
\vspace{-18pt}
\label{tab:harpo_ablations}
\end{table}

\textbf{Structured modulation.} We ablate the structured modulation via geometric reference, and instead calculate advantage modulation factors from the reciprocal of the contribution signals ($s^{(t)} = 1 / p^{(t)}$). From Tab.~\ref{tab:harpo_ablations}, this variant (w/o structured modulation) exhibits weaker multitask performance (avg. rank 2.00) compared to \names\ (1.90). An intuitive explanation is that the ablation produces a net amplification of advantages during training. The per-task modulation factors $s^{(t)}_{m}$ have a geometric mean that remains above 2.5 (App. Fig.~\ref{fig:appendix_add_factors}), indicating sustained advantage upscaling during training. In contrast, \names\ maintains modulation factors with geometric mean 1 due to geometric centering, mitigating uniform amplification of advantages across tasks. Since policy-gradient is scaled by advantage values, increasing advantage magnitudes can alter the effective update size which adversely impacts optimization~\citep{schulman2015trust, schulman2017proximal}. As a result, directly using $s^{(t)} = 1 / p^{(t)}$ instead of a structured, geometrically centered modulation may inadvertently impact performance.

% , potentially because the mean allocation factor $s^{(t)}$ is no longer constrained to $1$. 

\begin{figure}[t]
    \centering
    \includegraphics[width=\linewidth]{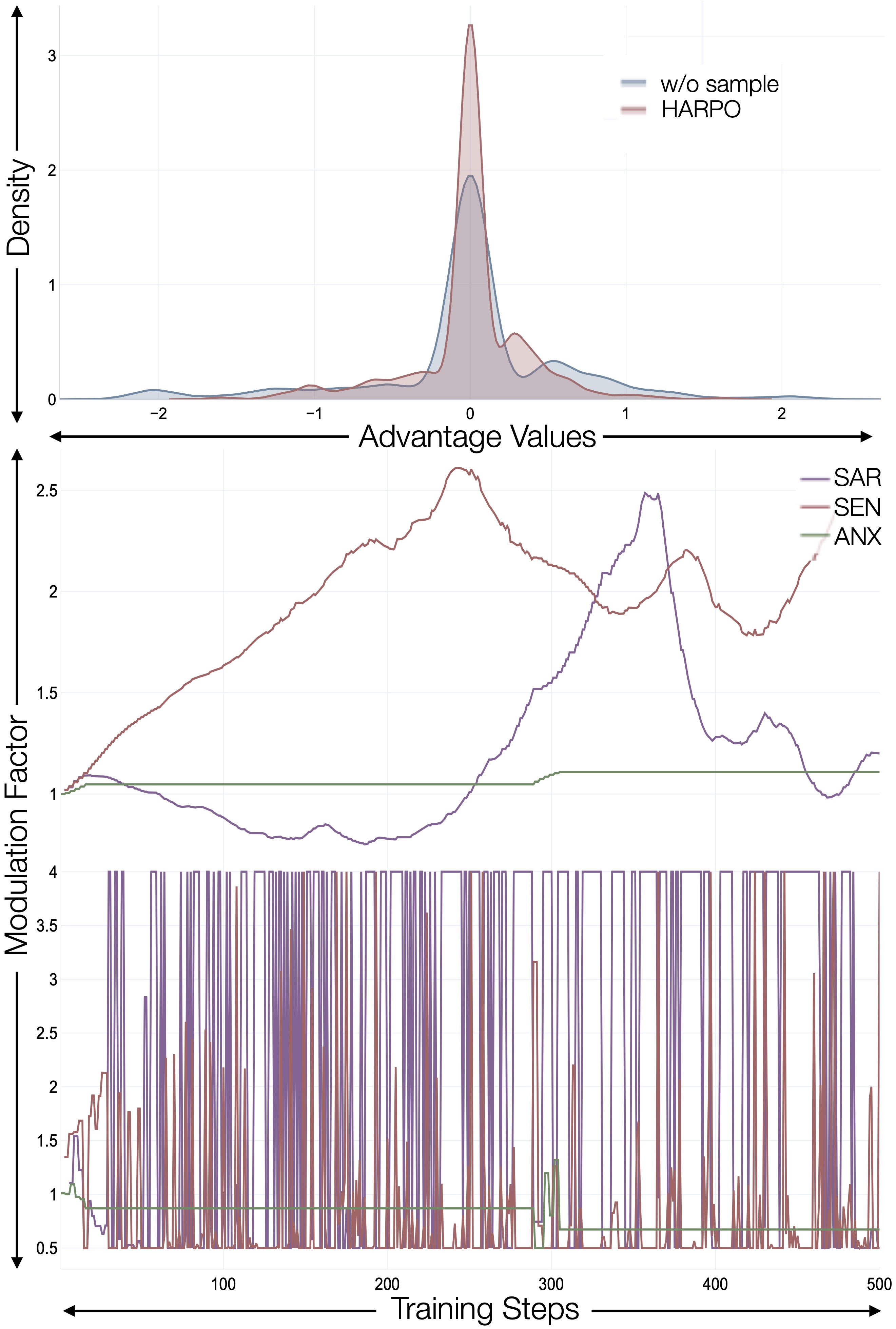}
    \vspace{-15pt} \caption{Top: Comparison of distribution of advantages between HARPO and the ablation without sample-level modulation, for the ANX task. Sample-level modulation leads to a narrowing of the advantage distribution which can result in better performance for specific tasks (additional task distributions are in App.~\ref{app:additional_results}). Middle \& Bottom: Comparison of modulation factors between HARPO (middle) and without inertial control (bottom). Inertial control enables a more gradual change in modulation factor values.}\vspace{-20pt}
\label{fig:hier_scaling_no_ema_ablations}
\end{figure}

\textbf{Inertial control.} From Tab.~\ref{tab:harpo_ablations}, the average rank degrades from 1.90 with the HARPO method to 2.70 for the ablation without inertial control (i.e., without smoothing in Eq.~\ref{eq:inertial_control}). To study this further, we examine the task modulation factors $s_{m}^{(t)}$ over training for tasks SAR, SEN, ANX. Tasks with greater performance improvements with inertial control, SAR (+7.96\%) and SEN (+5.69\%) from App.~\ref{app:additional_results} Tab.~\ref{tab:harpo_ablations_full}, also exhibit clearer differences in $s_{m}^{(t)}$ over training.  From Fig.~\ref{fig:hier_scaling_no_ema_ablations}, inertial control yields gradual changes in $s_{m}^{(t)}$, in contrast to sharper fluctuations without it. In contrast, ANX, which exhibits comparatively smaller performance differences (+1.30\%), shows similar $s_{m}^{(t)}$ trends with and without inertial control. This suggest that inertial control does not uniformly affect tasks, but can improve performance for tasks with excessive fluctuations in $s_{m}^{(t)}$, by stabilizing the scaling of advantage signals, which supports consistent policy updates~\citep{ilyas2018rlstable}.

% Modulation at the sample-level balances within-task learning by reducing rollout dominance.
\textbf{Modulation at the sample level.} To study modulating advantages at the sample-level in addition to the task-level, we run an ablation without sample-level modulation (i.e., leveraging only the task-level modulation factors, $s_{m}^{(t)}$). From Tab.~\ref{tab:harpo_ablations}, this results in weaker performance, with the ablation attaining a 2.60 avg. rank compared to \names\ at 1.90. To understand this, we analyze with Fig.~\ref{fig:hier_scaling_no_ema_ablations}, a task (ANX) that performs stronger with sample-level modulation. Accordingly, sample-level modulation produces a narrower advantage distribution than without, highlighting that relatively extreme advantage values are suppressed (we observe the same trend for other tasks NVC, SOC, HUM in App.~\ref{app:additional_plots} Fig.~\ref{fig:appendix_adv_dist}). This narrower spread does not reduce aggregate advantages, since the geometric mean enforced by Eqs.~\eqref{eq:geom_ref}-\eqref{eq:reciprocal_ratio} preserves the overall multiplicative scale of advantages. Instead, it prevents a subset of samples with extreme values from disproportionately influencing the policy update. 

% However, while beneficial for most tasks, the effects of sample-level modulation varies across tasks, as EMO, INT, and SOC show no explicit improvements (App.~\ref{app:full_ablation_results} Tab.~\ref{tab:harpo_ablations_full}).
% Empirically, we observe that this sustains training on these weaker tasks, as the mean rewards under \names\ steadily increases as compared to task-level only modulation (), which encounters stagnation.

% \paragraph{Controlling at the rollout level leads to higher performance overall.}


\begin{figure}[t]
  \centering
  \includegraphics[width=\columnwidth]{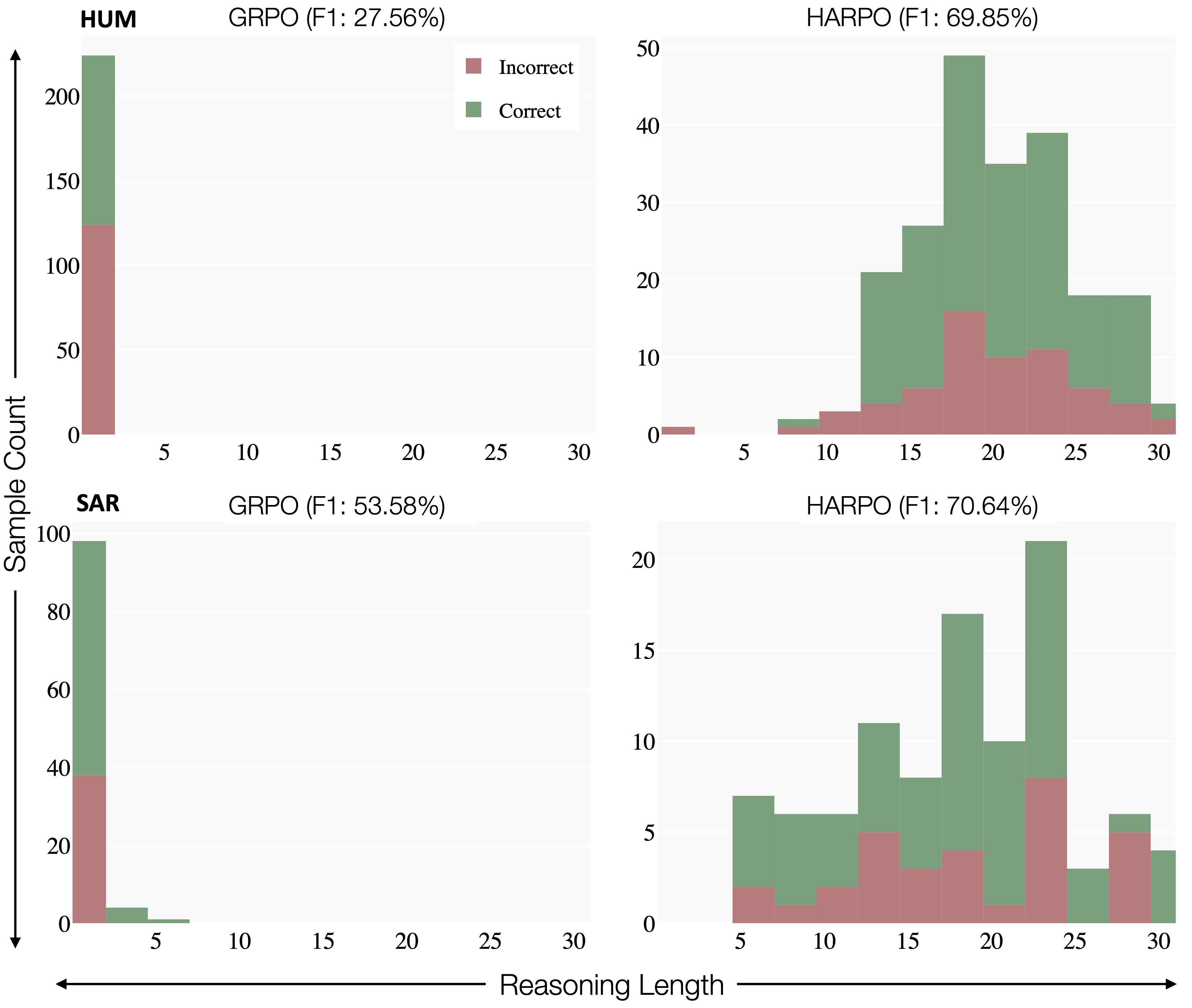}
    \vspace{-15pt} \caption{
  Sample count versus token reasoning length. Green indicates correct predictions, red indicates incorrect. HARPO induces more varied reasoning lengths for the respective tasks of (Top: HUM, Bottom: SAR), compared to GRPO.
  }
  \label{fig:reasoning_length_hist}\vspace{-10pt}
\end{figure}

\begin{figure}[h]
  \centering
  \includegraphics[width=\columnwidth]{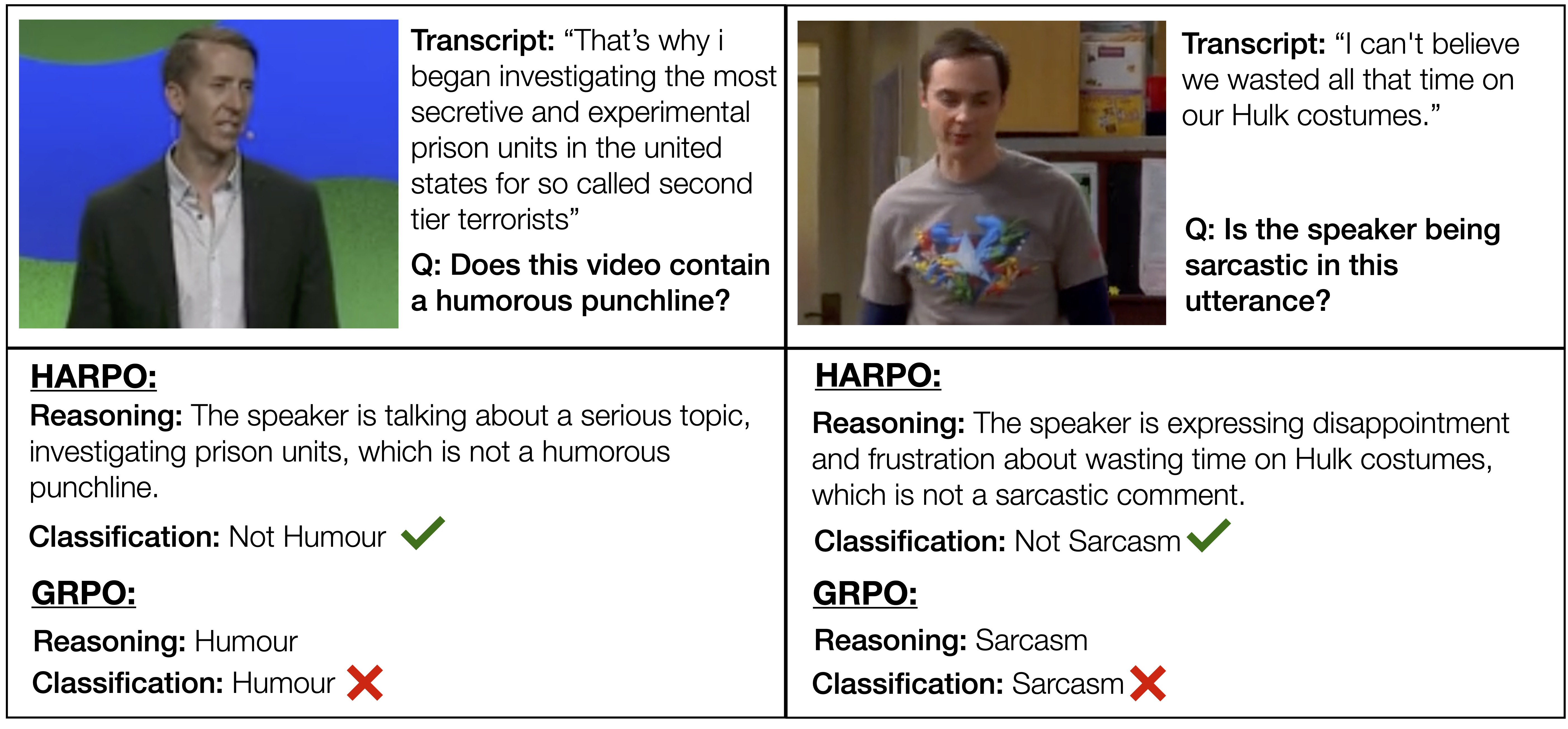}
  \vspace{-15pt}\caption{Example of reasoning traces on pragmatic tasks humor detection (HUM) (left) and sarcasm detection (SAR) (right). HARPO is observed to reflect more explicit and varied reasoning compared to GRPO, which defaults to minimal/ no reasoning.
  }
  \label{fig:reasoning_case_study_prag} \vspace{-15pt}
\end{figure}

\textbf{HARPO's effect on reasoning traces}. For complex pragmatic tasks such as sarcasm (SAR) and humor (HUM) detection, which rely on pragmatic cues (i.e., contextual signals beyond literal lexical content), we observe that HARPO induces richer reasoning behavior as compared to GRPO, at inference. From Fig.\ref{fig:reasoning_length_hist}, HARPO achieves a larger proportion of correct predictions on SAR and HUM, compared to GRPO. This is also accompanied by longer and more varied reasoning-length distributions. Qualitative analysis, Fig.~\ref{fig:reasoning_case_study_prag}, further illustrates that GRPO often produces shortcut responses with minimal or no explicit reasoning, while HARPO’s reasoning traces remain more contextually grounded, reflecting deliberation over pragmatic and inferential cues present in the input. These findings suggest that HARPO reduces the suppression of useful reasoning trajectories for these weaker tasks, resulting in richer reasoning behavior that coincides with improved task performance.

\vspace{-10pt}
\section{Conclusion}
%In this work, we introduce HARPO, a reinforcement learning method designed to account for heterogeneity in learning signals across multiple behavioral tasks. Using HARPO, we develop OmniSapiens 2.0, a unified behavioral model trained across diverse behavioral datasets. Our multitask and generalization evaluations show that OmniSapiens 2.0 achieves consistent performance across behavioral domains, highlighting the utility of accounting for heterogeneous learning signals.
In this work, we developed \namesm\, a unified foundation model for social behavior analysis trained using HARPO, a new RL method that accounts for heterogeneous learning signals across multiple behavioral tasks. Multitask and generalization evaluations showed significantly improved performance across behavioral tasks and increased interpretability of reasoning traces, underscoring the value of modeling heterogeneous learning signals through \names. More broadly, these findings suggest how social behavioral foundation models may benefit from explicitly protecting weaker, but structurally meaningful signals that encode latent social and behavioral structure. 

% We release \namesm\ and our accompanying code to facilitate future research by the community.

% To build the model, we introduced HARPO, a RL method that accounts for heterogeneous learning signals across multiple behavioral tasks. Using HARPO, we developed OmniSapiens 2.0, a unified behavioral model trained on diverse datasets. Multitask and generalization evaluations showed consistent performance across behavioral domains, underscoring the value of modeling heterogeneous learning signals. 

\section*{Acknowledgements}
This research/project is supported by the Asian Institute of Digital Finance, National University of Singapore; Ministry of Education, Singapore under its MOE Academic Research Fund Tier 2 (MOE-T2EP20123-0005: “Neurosymbolic AI for Commonsense-based Question Answering in Multiple Domains”), MOE Tier 1 Award (MOE-T2EP50221-0028: “Discipline-Informed Neural Networks for Interpretable Time-Series Discovery”), and by the RIE2025 Industry Alignment Fund – Industry Collaboration Projects (I2301E0026: “Generative AI”), administered by A*STAR, as well as supported by Alibaba Group and NTU Singapore. The authors would also like to acknowledge Modal and NVIDIA for their support through the provision of GPU and compute resources.

\section*{Impact Statement}
This paper presents work whose goal is to advance the field of 
Machine Learning. There are many potential societal consequences 
of our work, none of which we feel must be specifically highlighted here.


\bibliography{main}

@article{pessoa2008relationship,
	Author = {Pessoa, Luiz},
	Journal = {Nature reviews neuroscience},
	Number = {2},
	Pages = {148--158},
	Publisher = {Nature Publishing Group UK London},
	Title = {On the relationship between emotion and cognition},
	Volume = {9},
	Year = {2008}}

@article{ilyas2018rlstable,
	Author = {Ilyas, Andrew and Engstrom, Logan and Santurkar, Shibani and Tsipras, Dimitris and Janoos, Firdaus and Rudolph, Larry and Madry, Aleksander},
	Journal = {arXiv preprint arXiv:1811.02553},
	Title = {A closer look at deep policy gradients},
	Year = {2018}}

@conference{zadeh2018multimodal,
	Author = {Zadeh, Amir and Liang, Paul Pu and Poria, Soujanya and Cambria, Erik and Morency, Louis-Philippe},
	Booktitle = {ACL},
	Pages = {2236-2246},
	Title = {Multimodal Language Analysis in the Wild: {CMU-MOSEI} Dataset and Interpretable Dynamic Fusion Graph},
	Year = {2018}}

@conference{poria2018meld,
	Author = {Poria, Soujanya and Hazarika, Devamanyu and Majumder, Navonil and Naik, Gautam and Cambria, Erik and Mihalcea, Rada},
	Booktitle = {{ACL}},
	Pages = {527-536},
	Title = {{MELD: A} Multimodal Multi-Party Dataset for Emotion Recognition in Conversations},
	Year = {2019}}

@article{tess_dataset,
	Author = {Pichora-Fuller, M. Kathleen and Dupuis, Kate},
	Doi = {10.5683/SP2/E8H2MF},
	Journal = {Borealis},
	Title = {{Toronto emotional speech set (TESS)}},
	Url = {https://doi.org/10.5683/SP2/E8H2MF},
	Version = {V1},
	Year = {2020}}

@article{sawadogo2024ptsd,
	Author = {Sawadogo, Moctar Abdoul Latif and Pala, Furkan and Singh, Gurkirat and Selmi, Imen and Puteaux, Pauline and Othmani, Alice},
	Journal = {Multimedia Tools and Applications},
	Number = {14},
	Pages = {42861--42883},
	Publisher = {Springer},
	Title = {PTSD in the wild: a video database for studying post-traumatic stress disorder recognition in unconstrained environments},
	Volume = {83},
	Year = {2024}}

@article{cao2014crema,
	Author = {Cao, Houwei and Cooper, David G and Keutmann, Michael K and Gur, Ruben C and Nenkova, Ani and Verma, Ragini},
	Journal = {IEEE transactions on affective computing},
	Number = {4},
	Pages = {377--390},
	Publisher = {IEEE},
	Title = {Crema-d: Crowd-sourced emotional multimodal actors dataset},
	Volume = {5},
	Year = {2014}}

@article{kingma2014adam,
	Author = {Kingma, Diederik and Ba, Jimmy},
	Journal = {arXiv preprint arXiv:1412.6980},
	Title = {Adam: A method for stochastic optimization},
	Year = {2014}}

@article{zadeh2016mosi,
	Author = {Zadeh, Amir and Zellers, Rowan and Pincus, Eli and Morency, Louis-Philippe},
	Journal = {arXiv preprint arXiv:1606.06259},
	Title = {{MOSI}: Multimodal Corpus of Sentiment Intensity and Subjectivity Analysis in Online Opinion Videos},
	Year = {2016}}

@article{Busso2008IEMOCAP:Interactiveemotionaldyadic,
	Author = {Busso, Carlos and Bulut, Murtaza and Lee, Chi-Chun and Kazemzadeh, Abe and Mower, Emily and Kim, Samuel and Chang, Jeannette and Lee, Sungbok and Narayanan, Shrikanth S.},
	Bib2Html_Rescat = {emotion},
	Doi = {10.1007/s10579-008-9076-6},
	Journal = {Journal of Language Resources and Evaluation},
	Link = {http://sail.usc.edu/publications/files/BussoLRE2008.pdf},
	Month = {dec},
	Number = {4},
	Pages = {335-359},
	Title = {IEMOCAP: Interactive emotional dyadic motion capture database},
	Volume = {42},
	Year = {2008}}

@article{team2025gemma3,
	Author = {Team, Gemma and Kamath, Aishwarya and Ferret, Johan and Pathak, Shreya and Vieillard, Nino and Merhej, Ramona and Perrin, Sarah and Matejovicova, Tatiana and Ram{\'e}, Alexandre and Rivi{\`e}re, Morgane and others},
	Journal = {arXiv preprint arXiv:2503.19786},
	Title = {Gemma 3 technical report},
	Year = {2025}}

@article{bai2025qwen2.5vl,
	Author = {Bai, Shuai and Chen, Keqin and Liu, Xuejing and Wang, Jialin and Ge, Wenbin and Song, Sibo and Dang, Kai and Wang, Peng and Wang, Shijie and Tang, Jun and others},
	Journal = {arXiv preprint arXiv:2502.13923},
	Title = {Qwen2. 5-vl technical report},
	Year = {2025}}

@inproceedings{valstar2016avec,
	Author = {Valstar, Michel and Gratch, Jonathan and Schuller, Bj{\"o}rn and Ringeval, Fabien and Lalanne, Dennis and Torres Torres, Mercedes and Scherer, Stefan and Stratou, Giota and Cowie, Roddy and Pantic, Maja},
	Booktitle = {Proceedings of the 6th International Workshop on Audio/Visual Emotion Challenge},
	Organization = {ACM},
	Pages = {3--10},
	Title = {AVEC 2016: Depression, Mood, and Emotion Recognition Workshop and Challenge},
	Year = {2016}}

@inproceedings{castro2019towards,
	Author = {Castro, Santiago and Hazarika, Devamanyu and P{\'e}rez-Rosas, Ver{\'o}nica and Zimmermann, Roger and Mihalcea, Rada and Poria, Soujanya},
	Booktitle = {ACL},
	Title = {Towards Multimodal Sarcasm Detection (An \_Obviously\_ Perfect Paper)},
	Year = {2019}}

@inproceedings{hasan2019ur,
	Author = {Hasan, Md Kamrul and Rahman, Wasifur and Zadeh, AmirAli Bagher and Zhong, Jianyuan},
	Booktitle = {EMNLP-IJCNLP},
	Pages = {2046--2056},
	Title = {UR-FUNNY: A Multimodal Language Dataset for Understanding Humor},
	Year = {2019}}

@inproceedings{liang2018computational,
	Author = {Liang, Paul Pu and Salakhutdinov, Ruslan and Morency, Louis-Philippe},
	Booktitle = {ACL},
	Title = {Computational modeling of human multimodal language: The mosei dataset and interpretable dynamic fusion},
	Year = {2018}}

@article{xu2025qwen3,
	Author = {Xu, Jin and Guo, Zhifang and Hu, Hangrui and Chu, Yunfei and Wang, Xiong and He, Jinzheng and Wang, Yuxuan and Shi, Xian and He, Ting and Zhu, Xinfa and others},
	Journal = {arXiv preprint arXiv:2509.17765},
	Title = {Qwen3-omni technical report},
	Year = {2025}}

@article{liang2024foundations,
	Author = {Liang, Paul Pu and Zadeh, Amir and Morency, Louis-Philippe},
	Journal = {ACM Computing Surveys},
	Number = {10},
	Pages = {1--42},
	Publisher = {ACM New York, NY},
	Title = {Foundations \& trends in multimodal machine learning: Principles, challenges, and open questions},
	Volume = {56},
	Year = {2024}}

@inproceedings{yu2022chsimsv2,
	Author = {Yu, J. and others},
	Booktitle = {Proceedings of the 30th ACM International Conference on Multimedia},
	Pages = {1234--1243},
	Title = {CH-SIMS v2: A Chinese Multimodal Sentiment Analysis Dataset},
	Year = {2022}}

@inproceedings{zhang2025mmpsy,
	Author = {Zhang, X. and others},
	Booktitle = {Proceedings of the AAAI Conference on Artificial Intelligence},
	Title = {Mental-Perceiver: Multimodal Mental Health Dataset and Model},
	Year = {2025}}

@misc{siq2,
	Author = {Alex Wilf and Leena Mathur and Sheryl Mathew and Claire Ko and Youssouf Kebe and Paul Pu Liang and Louis-Philippe Morency},
	Howpublished = {\url{https://github.com/abwilf/Social-IQ-2.0-Challenge}},
	Journal = {GitHub repository},
	Publisher = {GitHub},
	Title = {Social-IQ 2.0 Challenge: Benchmarking Multimodal Social Understanding},
	Year = {2023}}

@inproceedings{li2023intentqa,
	Author = {Li, Jiapeng and Wei, Ping and Han, Wenjuan and Fan, Lifeng},
	Booktitle = {Proceedings of the IEEE/CVF international conference on computer vision},
	Pages = {11963--11974},
	Title = {Intentqa: Context-aware video intent reasoning},
	Year = {2023}}

@article{li2025mimeqa,
	Author = {Li, Hengzhi and Tjandrasuwita, Megan and Fung, Yi R and Solar-Lezama, Armando and Liang, Paul Pu},
	Journal = {arXiv preprint arXiv:2502.16671},
	Title = {Mimeqa: Towards socially-intelligent nonverbal foundation models},
	Year = {2025}}

@article{xu2025qwen2.5omni,
	Author = {Xu, Jin and Guo, Zhifang and He, Jinzheng and Hu, Hangrui and He, Ting and Bai, Shuai and Chen, Keqin and Wang, Jialin and Fan, Yang and Dang, Kai and others},
	Journal = {arXiv preprint arXiv:2503.20215},
	Title = {Qwen2. 5-omni technical report},
	Year = {2025}}

@misc{openai2025gpt5,
	Author = {{OpenAI}},
	Day = 7,
	Howpublished = {\url{https://openai.com/gpt-5/}},
	Month = aug,
	Title = {GPT-5},
	Year = {2025}}

@article{shao2024deepseekmath,
	Author = {Shao, Zhihong and Wang, Peiyi and Zhu, Qihao and Xu, Runxin and Song, Junxiao and Bi, Xiao and Zhang, Haowei and Zhang, Mingchuan and Li, YK and Wu, Yang and others},
	Journal = {arXiv preprint arXiv:2402.03300},
	Title = {Deepseekmath: Pushing the limits of mathematical reasoning in open language models},
	Year = {2024}}

@article{yu2025dapo,
	Author = {Yu, Qiying and Zhang, Zheng and Zhu, Ruofei and Yuan, Yufeng and Zuo, Xiaochen and Yue, Yu and Dai, Weinan and Fan, Tiantian and Liu, Gaohong and Liu, Lingjun and others},
	Journal = {arXiv preprint arXiv:2503.14476},
	Title = {Dapo: An open-source llm reinforcement learning system at scale},
	Year = {2025}}

@article{zhang2025grpo_lead,
	Author = {Zhang, Jixiao and Zuo, Chunsheng},
	Journal = {arXiv preprint arXiv:2504.09696},
	Title = {Grpo-lead: A difficulty-aware reinforcement learning approach for concise mathematical reasoning in language models},
	Year = {2025}}

@conference{ong2025human,
	Author = {Ong, Keane and Dai, Wei and Li, Carol and Feng, Dewei and Li, Hengzhi and Wu, Jingyao and Cheong, Jiaee and Mao, Rui and Mengaldo, Gianmarco and Cambria, Erik and others},
	Booktitle = {{Proceedings of ICLR}},
	Title = {Human behavior atlas: Benchmarking unified psychological and social behavior understanding},
	Year = {2026}}

@inproceedings{dang2023constrained,
	Author = {Dang, Ting and Dimitriadis, Antoni and Wu, Jingyao and Sethu, Vidhyasaharan and Ambikairajah, Eliathamby},
	Booktitle = {ICASSP 2023-2023 IEEE International Conference on Acoustics, Speech and Signal Processing (ICASSP)},
	Organization = {IEEE},
	Pages = {1--5},
	Title = {Constrained dynamical neural ode for time series modelling: A case study on continuous emotion prediction},
	Year = {2023}}

@article{giannakakis2019review,
	Author = {Giannakakis, Giorgos and Grigoriadis, Dimitris and Giannakaki, Katerina and Simantiraki, Olympia and Roniotis, Alexandros and Tsiknakis, Manolis},
	Journal = {IEEE transactions on affective computing},
	Number = {1},
	Pages = {440--460},
	Publisher = {IEEE},
	Title = {Review on psychological stress detection using biosignals},
	Volume = {13},
	Year = {2019}}

@article{joshi2022depression,
	Author = {Joshi, Manju Lata and Kanoongo, Nehal},
	Journal = {Materials Today: Proceedings},
	Pages = {217--226},
	Publisher = {Elsevier},
	Title = {Depression detection using emotional artificial intelligence and machine learning: A closer review},
	Volume = {58},
	Year = {2022}}

@article{miloyan2014future,
	Author = {Miloyan, Beyon and Pachana, Nancy A and Suddendorf, Thomas},
	Journal = {Cognition \& emotion},
	Number = {5},
	Pages = {795--810},
	Publisher = {Taylor \& Francis},
	Title = {The future is here: A review of foresight systems in anxiety and depression},
	Volume = {28},
	Year = {2014}}

@inproceedings{hessel2023androids,
	Author = {Hessel, Jack and Marasovi{\'c}, Ana and Hwang, Jena D and Lee, Lillian and Da, Jeff and Zellers, Rowan and Mankoff, Robert and Choi, Yejin},
	Booktitle = {Proceedings of {ACL}},
	Pages = {688--714},
	Title = {Do androids laugh at electric sheep? humor ``understanding'' benchmarks from the new yorker caption contest},
	Year = {2023}}

@article{monkaresi2016automated,
	Author = {Monkaresi, Hamed and Bosch, Nigel and Calvo, Rafael A and D'Mello, Sidney K},
	Journal = {IEEE Transactions on Affective Computing},
	Number = {1},
	Pages = {15--28},
	Publisher = {IEEE},
	Title = {Automated detection of engagement using video-based estimation of facial expressions and heart rate},
	Volume = {8},
	Year = {2016}}

@article{sutton1999policy,
	Author = {Sutton, Richard S and McAllester, David and Singh, Satinder and Mansour, Yishay},
	Journal = {Advances in neural information processing systems},
	Title = {Policy gradient methods for reinforcement learning with function approximation},
	Volume = {12},
	Year = {1999}}

@article{yang2025humanomniv2,
	Author = {Yang, Qize and Yao, Shimin and Chen, Weixuan and Fu, Shenghao and Bai, Detao and Zhao, Jiaxing and Sun, Boyuan and Yin, Bowen and Wei, Xihan and Zhou, Jingren},
	Journal = {arXiv preprint arXiv:2506.21277},
	Title = {HumanOmniV2: From Understanding to Omni-Modal Reasoning with Context},
	Year = {2025}}

@article{arora2012multiplicative,
	Author = {Arora, Sanjeev and Hazan, Elad and Kale, Satyen},
	Journal = {Theory of computing},
	Number = {1},
	Pages = {121--164},
	Publisher = {Theory of Computing Exchange},
	Title = {The multiplicative weights update method: a meta-algorithm and applications},
	Volume = {8},
	Year = {2012}}

@article{bubeck2015convex,
	Author = {Bubeck, S{\'e}bastien and others},
	Journal = {Foundations and Trends{\textregistered} in Machine Learning},
	Number = {3-4},
	Pages = {231--357},
	Publisher = {Now Publishers, Inc.},
	Title = {Convex optimization: Algorithms and complexity},
	Volume = {8},
	Year = {2015}}

@article{teh2017distral,
	Author = {Teh, Yee and Bapst, Victor and Czarnecki, Wojciech M and Quan, John and Kirkpatrick, James and Hadsell, Raia and Heess, Nicolas and Pascanu, Razvan},
	Journal = {Advances in neural information processing systems},
	Title = {Distral: Robust multitask reinforcement learning},
	Volume = {30},
	Year = {2017}}

@inproceedings{schulman2015trust,
	Author = {Schulman, John and Levine, Sergey and Abbeel, Pieter and Jordan, Michael and Moritz, Philipp},
	Booktitle = {International conference on machine learning},
	Organization = {PMLR},
	Pages = {1889--1897},
	Title = {Trust region policy optimization},
	Year = {2015}}

@inproceedings{henderson2018deep,
	Author = {Henderson, Peter and Islam, Riashat and Bachman, Philip and Pineau, Joelle and Precup, Doina and Meger, David},
	Booktitle = {Proceedings of the AAAI conference on artificial intelligence},
	Number = {1},
	Title = {Deep reinforcement learning that matters},
	Volume = {32},
	Year = {2018}}

@article{chu2025gpg,
	Author = {Chu, Xiangxiang and Huang, Hailang and Zhang, Xiao and Wei, Fei and Wang, Yong},
	Journal = {arXiv preprint arXiv:2504.02546},
	Title = {Gpg: A simple and strong reinforcement learning baseline for model reasoning},
	Year = {2025}}

@article{hu2025reinforce++,
	Author = {Hu, Jian},
	Journal = {arXiv preprint arXiv:2501.03262},
	Title = {Reinforce++: A simple and efficient approach for aligning large language models},
	Year = {2025}}

@article{ahmadian2024rloo,
	Author = {Ahmadian, Arash and Cremer, Chris and Gall{\'e}, Matthias and Fadaee, Marzieh and Kreutzer, Julia and Pietquin, Olivier and {\"U}st{\"u}n, Ahmet and Hooker, Sara},
	Journal = {arXiv preprint arXiv:2402.14740},
	Title = {Back to basics: Revisiting reinforce style optimization for learning from human feedback in llms},
	Year = {2024}}

@article{schulman2017proximal,
	Author = {Schulman, John and Wolski, Filip and Dhariwal, Prafulla and Radford, Alec and Klimov, Oleg},
	Journal = {arXiv preprint arXiv:1707.06347},
	Title = {Proximal policy optimization algorithms},
	Year = {2017}}

@article{deng2024avasd,
	Author = {Deng, Shijian and Kosloski, Erin E and Patel, Siddhi and Barnett, Zeke A and Nan, Yiyang and Kaplan, Alexander and Aarukapalli, Sisira and Doan, William T and Wang, Matthew and Singh, Harsh and others},
	Journal = {IEEE Transactions on Multimedia},
	Publisher = {IEEE},
	Title = {Hear me, see me, understand me: Audio-visual autism behavior recognition},
	Year = {2024}}

@inproceedings{yu2020gradient,
	Author = {Yu, Tianhe and Kumar, Saurabh and Gupta, Abhishek and Levine, Sergey and Hausman, Karol and Finn, Chelsea},
	Booktitle = {Advances in Neural Information Processing Systems},
	Pages = {5824--5836},
	Title = {Gradient Surgery for Multi-Task Learning},
	Volume = {33},
	Year = {2020}}

@article{zhang2025surveyreasoning,
	Author = {Zhang, Kaiyan and Zuo, Yuxin and He, Bingxiang and Sun, Youbang and Liu, Runze and Jiang, Che and Fan, Yuchen and Tian, Kai and Jia, Guoli and Li, Pengfei and others},
	Journal = {arXiv preprint arXiv:2509.08827},
	Title = {A survey of reinforcement learning for large reasoning models},
	Year = {2025}}

@inproceedings{kendall2018multi,
	Author = {Kendall, Alex and Gal, Yarin and Cipolla, Roberto},
	Booktitle = {Proceedings of the IEEE Conference on Computer Vision and Pattern Recognition},
	Pages = {7482--7491},
	Title = {Multi-Task Learning Using Uncertainty to Weigh Losses for Scene Geometry and Semantics},
	Year = {2018}}

@article{baron2001intercontheory,
	Author = {Baron-Cohen, Simon},
	Journal = {Prisme},
	Number = {1},
	Pages = {74--183},
	Title = {Theory of mind in normal development and autism},
	Volume = {34},
	Year = {2001}}

@book{goleman2006social,
	Author = {Goleman, Daniel},
	Publisher = {Bantam},
	Title = {Social intelligence: The new science of human relationships},
	Year = {2006}}

@article{kihlstrom2000social,
	Author = {Kihlstrom, John F and Cantor, Nancy},
	Publisher = {Cambridge University Press},
	Title = {Social intelligence.},
	Year = {2000}}

@article{feng2025emagrpo,
	Author = {Feng, Kaituo and Zhang, Manyuan and Li, Hongyu and Fan, Kaixuan and Chen, Shuang and Jiang, Yilei and Zheng, Dian and Sun, Peiwen and Zhang, Yiyuan and Sun, Haoze and others},
	Journal = {arXiv preprint arXiv:2512.03043},
	Title = {Onethinker: All-in-one reasoning model for image and video},
	Year = {2025}}

@article{langer2021xai,
	Author = {Langer, Markus and Oster, Daniel and Speith, Timo and Hermanns, Holger and K{\"a}stner, Lena and Schmidt, Eva and Sesing, Andreas and Baum, Kevin},
	Journal = {Artificial intelligence},
	Pages = {103473},
	Publisher = {Elsevier},
	Title = {What do we want from Explainable Artificial Intelligence (XAI)?--A stakeholder perspective on XAI and a conceptual model guiding interdisciplinary XAI research},
	Volume = {296},
	Year = {2021}}

@article{nauta2023anecdotalxai,
	Author = {Nauta, Meike and Trienes, Jan and Pathak, Shreyasi and Nguyen, Elisa and Peters, Michelle and Schmitt, Yasmin and Schl{\"o}tterer, J{\"o}rg and Van Keulen, Maurice and Seifert, Christin},
	Journal = {ACM Computing Surveys},
	Number = {13s},
	Pages = {1--42},
	Publisher = {ACM New York, NY},
	Title = {From anecdotal evidence to quantitative evaluation methods: A systematic review on evaluating explainable ai},
	Volume = {55},
	Year = {2023}}

@article{alvarez2018towardsinterp,
	Author = {Alvarez Melis, David and Jaakkola, Tommi},
	Journal = {Advances in neural information processing systems},
	Title = {Towards robust interpretability with self-explaining neural networks},
	Volume = {31},
	Year = {2018}}

@article{wang2022self,
	Author = {Wang, Xuezhi and Wei, Jason and Schuurmans, Dale and Le, Quoc and Chi, Ed and Narang, Sharan and Chowdhery, Aakanksha and Zhou, Denny},
	Journal = {arXiv preprint arXiv:2203.11171},
	Title = {Self-consistency improves chain of thought reasoning in language models},
	Year = {2022}}

@inproceedings{jie2024interpretable,
	Author = {Jie, Yeo Wei and Satapathy, Ranjan and Goh, Rick and Cambria, Erik},
	Booktitle = {Findings of the Association for Computational Linguistics: NAACL 2024},
	Pages = {2148--2164},
	Title = {How interpretable are reasoning explanations from prompting large language models?},
	Year = {2024}}

@inproceedings{shen2022automaticeatd,
	Author = {Shen, Ying and Yang, Huiyu and Lin, Lin},
	Booktitle = {ICASSP 2022-2022 IEEE International Conference on Acoustics, Speech and Signal Processing (ICASSP)},
	Organization = {IEEE},
	Pages = {6247--6251},
	Title = {Automatic depression detection: An emotional audio-textual corpus and a gru/bilstm-based model},
	Year = {2022}}

@inproceedings{niu2016sentimentmvsd,
	Author = {Niu, Teng and Zhu, Shiai and Pang, Lei and El Saddik, Abdulmotaleb},
	Booktitle = {International conference on multimedia modeling},
	Organization = {Springer},
	Pages = {15--27},
	Title = {Sentiment analysis on multi-view social data},
	Year = {2016}}

@inproceedings{yue2024sarcnet,
	Author = {Yue, Tan and Shi, Xuzhao and Mao, Rui and Hu, Zonghai and Cambria, Erik},
	Booktitle = {Proceedings of {LREC-COLING}},
	Pages = {14325--14335},
	Title = {{SarcNet: A} multilingual multimodal sarcasm detection dataset},
	Year = {2024}}

@article{wei2024revisiting,
	Author = {Wei, Jiawen and Turb{\'e}, Hugues and Mengaldo, Gianmarco},
	Journal = {arXiv preprint arXiv:2407.19683},
	Title = {Revisiting the robustness of post-hoc interpretability methods},
	Year = {2024}}

@article{turbe2023evaluation,
	Author = {Turb{\'e}, Hugues and Bjelogrlic, Mina and Lovis, Christian and Mengaldo, Gianmarco},
	Journal = {Nature Machine Intelligence},
	Number = {3},
	Pages = {250--260},
	Publisher = {Nature Publishing Group UK London},
	Title = {Evaluation of post-hoc interpretability methods in time-series classification},
	Volume = {5},
	Year = {2023}}

@article{ong2025explainable,
  title={Explainable natural language processing for corporate sustainability analysis},
  author={Ong, Keane and Mao, Rui and Satapathy, Ranjan and Shirota Filho, Ricardo and Cambria, Erik and Sulaeman, Johan and Mengaldo, Gianmarco},
  journal={Information Fusion},
  volume={115},
  pages={102726},
  year={2025},
  publisher={Elsevier}
}
\bibliographystyle{icml2025}

%%%%%%%%%%%%%%%%%%%%%%%%%%%%%%%%%%%%%%%%%%%%%%%%%%%%%%%%%%%%%%%%%%%%%%%%%%%%%%%
%%%%%%%%%%%%%%%%%%%%%%%%%%%%%%%%%%%%%%%%%%%%%%%%%%%%%%%%%%%%%%%%%%%%%%%%%%%%%%%
% APPENDIX
%%%%%%%%%%%%%%%%%%%%%%%%%%%%%%%%%%%%%%%%%%%%%%%%%%%%%%%%%%%%%%%%%%%%%%%%%%%%%%%
%%%%%%%%%%%%%%%%%%%%%%%%%%%%%%%%%%%%%%%%%%%%%%%%%%%%%%%%%%%%%%%%%%%%%%%%%%%%%%%
\newpage
\appendix
\onecolumn

\section{HARPO Algorithm Block}\label{app:algo}
We describe the full HARPO procedure in Algorithm~\ref{alg:harpo_full}. Unlike Eq.~\eqref{eq:inertial_control}, where notation is collapsed for simplicity, the full algorithm explicitly distinguishes instantaneous contribution signals and modulation factors of the rollout batch, $\{p^{(t)}, s^{(t)}\}$, from their inertially controlled running estimates, $\{\tilde p^{(t)}, \tilde s^{(t)}\}$. The latter serves as the operational quantities used in modulation.

\begin{algorithm}[h!]
\small
   \caption{\names: Heterogeneity Aware Relative Policy Optimization. }
   \label{alg:harpo_full}
   \begin{minipage}{\linewidth}
   \begin{algorithmic}
   \STATE {\bfseries Input:} task set $\mathcal M$; batch $\{(m,q)\}\sim\mathcal D$, where $m\in\mathcal M$ indexes tasks and $q$ denotes a sample from task $m$;
   rollout groups $\{G_{(m,q)}\}$ with rewards $\{r_{(m,q,i)}\}$; $\varepsilon>0$; $\beta_\rho,\beta_s\in[0,1)$; previous inertial states $\tilde p^{(t-1)}$ and $\tilde s^{(t-1)}$
   \STATE {\bfseries Output:} updated policy parameters $\theta$

   \STATE \textbf{Construct group-normalized advantages:}
   \FORALL{$(m,q)$ in batch}
      \STATE $\hat\mu_{(m,q)} \leftarrow \frac{1}{|G_{(m,q)}|}\sum_{i\in G_{(m,q)}} r_{(m,q,i)}$
      \STATE $\hat\sigma_{(m,q)} \leftarrow \sqrt{\frac{1}{|G_{(m,q)}|}\sum_{i\in G_{(m,q)}}(r_{(m,q,i)}-\hat\mu_{(m,q)})^2}$
      \FORALL{$i\in G_{(m,q)}$}
         \STATE $\hat A^{(t)}_{(m,q,i)} \leftarrow \dfrac{r_{(m,q,i)}-\hat\mu_{(m,q)}}{\hat\sigma_{(m,q)}+\varepsilon}$
      \ENDFOR
   \ENDFOR

   \STATE \textbf{Construct contribution signals:}
   \FORALL{$(m,q)$ in batch}
      \STATE $p^{(t)}_{(m,q)} \leftarrow \frac{1}{|G_{(m,q)}|}\sum_{i\in G_{(m,q)}}\left|\hat A^{(t)}_{(m,q,i)}\right|$
      \STATE $\tilde p^{(t)}_{(m,q)} \leftarrow \beta_\rho\,\tilde p^{(t-1)}_{(m,q)} + (1-\beta_\rho)\,p^{(t)}_{(m,q)}$
   \ENDFOR
   \FORALL{$m$ with samples in batch}
      \STATE $p^{(t)}_{m} \leftarrow
      \dfrac{\sum_{q\in \mathcal Q_m^{(t)}} \sum_{i\in G_{(m,q)}} \left|\hat A^{(t)}_{(m,q,i)}\right|}
            {\sum_{q\in \mathcal Q_m^{(t)}} |G_{(m,q)}|}$
      \STATE $\tilde p^{(t)}_{m} \leftarrow \beta_\rho\,\tilde p^{(t-1)}_{m} + (1-\beta_\rho)\,p^{(t)}_{m}$
   \ENDFOR

   \STATE \textbf{Construct geometric references:}
   \FORALL{$m$ with samples in batch}
      \STATE $\bar p^{(t)}_{\mathrm{ref},m} \leftarrow \Big(\prod_{q\in\mathcal Q_m^{(t)}} \tilde p^{(t)}_{(m,q)}\Big)^{1/|\mathcal Q_m^{(t)}|}$
   \ENDFOR
   \STATE $\bar p^{(t)}_{\mathrm{ref},\mathcal M} \leftarrow \Big(\prod_{m\in\mathcal M} \tilde p^{(t)}_{m}\Big)^{1/|\mathcal M|}$

   \STATE \textbf{Construct modulation factors:}
   \FORALL{$(m,q)$ in batch}
      \STATE $s^{(t)}_{(m,q)} \leftarrow \bar p^{(t)}_{\mathrm{ref},m} / \tilde p^{(t)}_{(m,q)}$
      \STATE $\tilde s^{(t)}_{(m,q)} \leftarrow \left(\tilde s^{(t-1)}_{(m,q)}\right)^{\beta_s}\left(s^{(t)}_{(m,q)}\right)^{1-\beta_s}$
   \ENDFOR
   \FORALL{$m$ with samples in batch}
      \STATE $s^{(t)}_{m} \leftarrow \bar p^{(t)}_{\mathrm{ref},\mathcal M} / \tilde p^{(t)}_{m}$
      \STATE $\tilde s^{(t)}_{m} \leftarrow \left(\tilde s^{(t-1)}_{m}\right)^{\beta_s}\left(s^{(t)}_{m}\right)^{1-\beta_s}$
   \ENDFOR

   \STATE \textbf{Obtain HARPO-modulated advantages:}
   \FORALL{$(m,q)$ in batch}
      \FORALL{$i\in G_{(m,q)}$}
         \STATE $A^{\mathrm{H}}{}^{(t)}_{(m,q,i)} \leftarrow \tilde s^{(t)}_{(m,q)}\, \tilde s^{(t)}_{m}\, \hat A^{(t)}_{(m,q,i)}$
      \ENDFOR
   \ENDFOR

   \STATE \textbf{Optimize policy with HARPO objective:}
   \STATE Construct PPO-clipped surrogate $\tilde A^{\mathrm{H}}_{(m,q,i):k}(\theta)$ from $A^{\mathrm{H}}{}^{(t)}_{(m,q,i)}$, in the same fashion as Eq.~\eqref{eq:grpo_compact2}
   \STATE Update $\theta$ by maximizing $J_{\names}(\theta)$ using $\tilde A^{\mathrm{H}}_{(m,q,i):k}(\theta)$  (Eq.~\eqref{eq:harpo_compact_controlled})
   \end{algorithmic}
   \end{minipage}
\end{algorithm}

\clearpage
\section{Additional Details on the Human Behavioral Atlas Benchmark}

\begin{figure*}[h]
  \centering
\includegraphics[width=\textwidth]{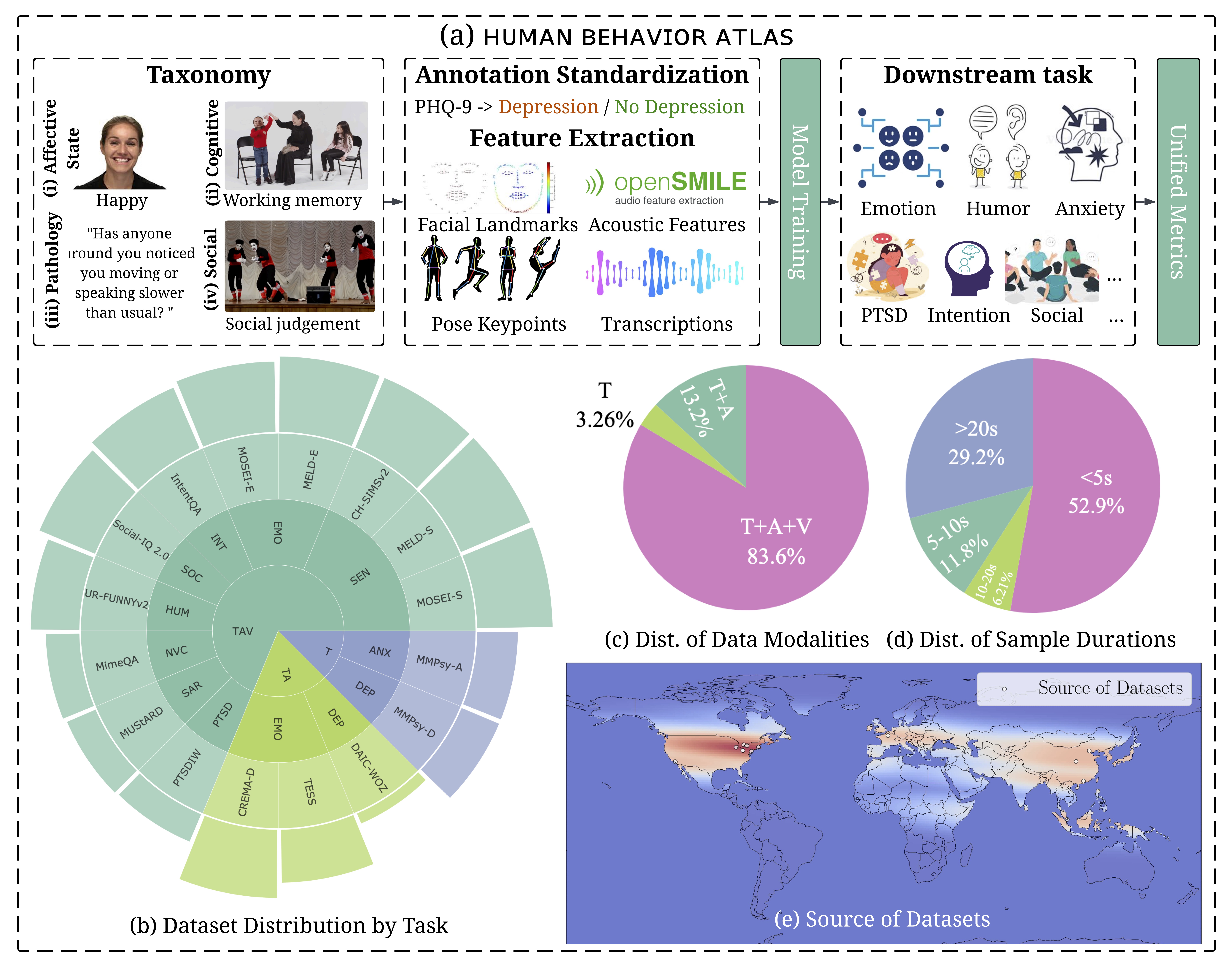}
\vspace{-20pt}
\caption{Overview of the Human Behavior Atlas (HBA) benchmark~\cite{ong2025human}, a large-scale benchmark for unified multimodal social behavior understanding and reasoning. HBA comprises diverse datasets spanning affective, cognitive, pathological, and social behavioral analysis tasks across text, audio, and video modalities. (a) Dataset selection criteria and preprocessing pipeline. (b) Distribution of datasets across 10 behavioral tasks and multimodal input settings. (c) Distribution of data modalities, highlighting the benchmark's strong emphasis on video-centric multimodal understanding. (d) Distribution of sample durations across short- and long-range behavioral understanding scenarios. (e) Geographic distribution of dataset sources across North America, Europe, and Asia.}
  \label{fig:hba_overview}
\end{figure*}

\subsection{Tasks and Datasets}

Human Behavioral Atlas~\citep{ong2025human} comprises approximately 100k samples, with 10 tasks and 13 datasets; we summarize the benchmark below for completeness, with details following the original benchmark description:

The tasks for the benchmark include Sentiment polarity (SEN): classifying attitudes as positive, negative, or neutral; emotion recognition (EMO): identifying emotions (anger, joy, sadness); social reasoning (SOC): understanding socially grounded judgments like empathy or appropriateness; intent recognition (INT): identifying the underlying purpose behind a behavior; and non-verbal communication (NVC): interpreting gestures and facial expressions. They also cover humor detection (HUM), sarcasm detection (SAR), anxiety detection (ANX), depression detection (DEP), and PTSD detection (PTSD).

Each task may be associated with one or more datasets. We summarize the benchmark datasets as follows. CMU-MOSEI~\cite{zadeh2018multimodal} is a large-scale multimodal dataset annotated for sentiment and emotion in real-world opinionated videos. MELD~\cite{poria2018meld} provides utterance-level emotion and sentiment annotations from multi-party dialogues in the TV series \textit{Friends}. UR-FUNNYv2~\cite{hasan2019ur} contains multimodal TED talk clips annotated for humor. MUStARD~\cite{castro2019towards} is a multimodal sarcasm detection dataset constructed from television show dialogues. DAIC-WOZ~\cite{valstar2016avec} consists of multimodal clinical interviews for depression assessment. CREMA-D~\cite{cao2014crema} is an acted emotional speech dataset with categorical emotion labels. CH-SIMSv2~\cite{yu2022chsimsv2} comprises multimodal sentiment annotations from simulated human–computer interactions. MMPsy~\cite{zhang2025mmpsy} is a multimodal mental health dataset annotated for anxiety and depression. PTSD in the Wild~\cite{sawadogo2024ptsd} contains real-world videos annotated for post-traumatic stress disorder. TESS~\cite{tess_dataset} is an emotional speech dataset with acted portrayals of discrete emotions. Social-IQ 2~\cite{siq2} evaluates social intelligence via reasoning over human interactions in video. IntentQA~\cite{li2023intentqa} is a video question-answering dataset focused on intent inference in everyday scenarios. MimeQA~\cite{li2025mimeqa} evaluates nonverbal social reasoning using gesture-based video question answering.

We report from the original paper, a Tab.~\ref{tab:datasets_summarised} which summarizes the datasets associated task, modalities, sample count and evaluation metric. For completeness, the paper also includes the different dimensions of human behavior that each dataset belongs to. These include, \emph{affective states} (Aff), which capture feelings, emotions and sentiments; \emph{cognitive states} (Cog), which reflect internal mental processes such as reasoning or attention inferred from observable behavior; \emph{pathological states} (Path), which correspond to psychological or psychiatric conditions assessed through verbal or nonverbal indicators; and \emph{social processes} (Soc), which characterize social interaction and communicative behaviors such as humor, intent, and cooperation.

\begin{table*}[h]
\centering
\caption{Datasets and their associated tasks and human behavior dimensions in Human Behavior Atlas. The modalities T, A, V stand for text, audio and vision respectively. The tasks can fall into two categories, CLS = classification (evaluated by direct label matching). TXTR = text-response (evaluated by an LLM judge).}
\label{tab:datasets_summarised}
\small
\renewcommand{\arraystretch}{1.1}
\setlength{\tabcolsep}{4pt}
\begin{tabular}{l p{2.2cm} p{2.0cm} c c c p{4.8cm}}
\toprule
\textbf{Dataset} & \textbf{Dimension} & \textbf{Task(s)} & \textbf{Task Type} & \textbf{Modalities} & \textbf{Samples} & \textbf{Eval. Metric} \\
\midrule
CMU-MOSEI & Aff; Cog & EMO, SEN & CLS & T / A / V & 31,454 & Binary weighted F1 (SEN), Mean weighted acc. (EMO) \\
MELD & Aff; Soc; Cog & EMO, SEN & CLS & T / A / V & 27,412 & Binary weighted F1 (SEN), Mean weighted acc. (EMO) \\
TESS & Aff; Cog & EMO & CLS & T / A / -- & 2,800 & Mean weighted accuracy \\
CREMA\textendash D & Aff & EMO & CLS & T / A / -- & 7,442 & Mean weighted accuracy \\
CH\textendash SIMSv2 & Aff & SEN & CLS & T / A / V & 4,403 & Binary weighted F1 \\
Social-IQ 2.0 & Soc; Cog & SOC & TXTR & T / A / V & 6,437 & Accuracy (LLM\textendash Judge) \\
IntentQA & Soc; Cog & INT & TXTR & T / A / V & 16,297 & Accuracy (LLM\textendash Judge) \\
MimeQA & Soc & NVC & TXTR & T / A / V & 806 & Accuracy (LLM\textendash Judge) \\
UR\textendash FUNNYv2 & Soc & HUM & CLS & T / A / V & 2,125 & Weighted F1 \\
MUStARD & Soc & SAR & CLS & T / A / V & 690 & Weighted F1 \\
DAIC\textendash WOZ & Path & DEP & CLS & T / A / -- & 189 & Weighted F1 \\
MMPsy & Path & DEP, ANX & CLS & T / -- / -- & 1,275 & Weighted F1 \\
PTSD\textendash in\textendash the\textendash Wild & Path & PTSD & CLS & T / A / V & 634 & Weighted F1 \\
\bottomrule
\end{tabular}
\end{table*}

\subsection{Evaluation Metrics}
We follow the same evaluation metrics utilized in Human Behavior Atlas, which specifically account for the task-specific nuances. The summary of the metrics utilized for each task is found in Tab.~\ref{tab:datasets_summarised}.

For HUM (Humor Detection), SAR (Sarcasm Detection), DEP (Depression Detection), ANX (Anxiety Detection), and PTSD (PTSD Detection), the weighted F1 score is computed:  
\[
F1 = \frac{2 \cdot \text{Precision} \cdot \text{Recall}}{\text{Precision} + \text{Recall}},
\]  
where,  
\[
\text{Precision} = \frac{TP}{TP + FP}, 
\qquad 
\text{Recall} = \frac{TP}{TP + FN}.
\]  
The weighted F1 is then computed as:  
\[
\text{Weighted-F1} = \sum_{c \in C} \frac{n_c}{N} \cdot F1_c,
\]  
where $n_c$ is the number of true instances in class $c$, $N$ is the total number of instances, and $C$ is the set of classes.  

SEN (Sentiment Detection) utilizes binary weighted F1, which applies the same formula but only over the positive and negative sentiment classes, where fine-grained sentiment scales (i.e., weakly positive or weakly negative) are mapped into positive or negative classes respectively. This accounts for the differences in sentiment-scale labelling across the SEN datasets. 

For EMO (Emotion Recognition), the mean/ average weighted accuracy across all emotion classes (e.g., fear, surprise, joy) is calculated, using the weighted accuracy formula, following~\citet{liang2018computational}:  
\[
\text{Weighted-Accuracy} = 0.5 \cdot \frac{TP}{P} + 0.5 \cdot \frac{TN}{N},
\]  
where $TP$ and $TN$ are the number of true positives and true negatives for the target class, and $P$ and $N$ denote the total number of positive and negative samples, respectively.

For free-text response QA tasks such as NVC (Non-Verbal Communication), INT (Intent Recognition), and SOC (Social Reasoning), an LLM judge (GPT-5 nano~\citep{openai2025gpt5}) is used to grade the generated responses. Specifically, task-specific prompts are provided to the LLM judge and the proportion of responses marked as TRUE is recorded as an estimate of accuracy:  
\[
\text{Accuracy} = \frac{\text{number of TRUE responses}}{\text{number of total responses}}
\] 

The full grading prompts can be found in the original Human Behavior Atlas paper~\citep{ong2025human}.

\section{Experimental Settings.}
\subsection{Hyperparameter Settings}
We train all methods for up to 5 epochs with early stopping based on validation performance. Validation is performed every 50 training steps. Early stopping is triggered if the validation metric does not improve for 5 consecutive validation runs (i.e., over 250
training steps). This stopping criterion helps reduce the impact of short-horizon fluctuations in validation metrics that arise from the high variance and non-stationarity of on-policy sampling~\citep{henderson2018deep}. We select the checkpoint achieving the highest average validation weighted F1 score.

From Tab.~\ref{tab:models_and_algos_multi_task_results}, model results (i.e., Gemma-3-4B~\citep{team2025gemma3}, Qwen 2.5-Omni-7B~\citep{xu2025qwen2.5omni}, Qwen 2.5-VL-7B~\citep{bai2025qwen2.5vl}, OmniSapiens-7B RL~\citep{ong2025human}, HumanOmniV2-7B~\citep{yang2025humanomniv2} are taken from the Human Behavior Atlas benchmark paper~\citep{ong2025human}), while we run the evaluation of Qwen 3-VL-8B Instruct~\citep{xu2025qwen3} using its implementation on Huggingface\footnote{https://huggingface.co}.

On the other hand, we implement the reinforcement learning training algorithms in Tab.~\ref{tab:models_and_algos_multi_task_results} (i.e., RLOO~\citep{ahmadian2024rloo}, RE++~\citep{hu2025reinforce++}, GPG~\citep{chu2025gpg}, GRPO~\citep{shao2024deepseekmath}) using the VERL package~\footnote{https://github.com/volcengine/verl}. To enable fair comparison with HARPO, all reinforcement learning methods are run on the same Human Behavior Atlas benchmark, with the same base model, Qwen 2.5-Omni-7B~\citep{xu2025qwen2.5omni}, and exactly the same reward design in Sec.~\ref{sec:reward_design}. For the RL baselines, we follow standard practice by reusing the hyperparameter configurations reported in the original papers, thereby preserving the authors' intended optimization settings. For HARPO, we retain the learning rate used by GRPO ($1\times10^{-6}$), since HARPO retains the update structure as GRPO, and we set $\beta_\rho$, $\beta_s$ in the inertial control Eq.~\eqref{eq:inertial_control} as 0.95. We also utilize the AdamW optimizer and omit explicit KL regularization. This is motivated by HARPO’s advantage modulation, which already acts to regularize policy updates, and is aligned with prior literature that strict KL constraints are not always necessary in on-policy training~\citep{yu2025dapo}. For all reinforcement learning methods, we fix the number of rollouts to 5, utilize the same effective batch size of 256 with PPO-mini-batch size of 128, and set the prompt and response lengths to 4096 and 2048 respectively. All methods are run on 4 Nvidia H200s as well as 4 Nvidia RTX PRO 6000 Blackwell GPUs.

\subsection{Full Details on Reward Design}~\label{app:reward_full_formulation}

From Sec.~\ref{sec:reward_design}, we utilize a final reward per sample that combines task accuracy $r_{task}$, format correcteness $r_{fmt}$, length penalty $r_{len}$. We summarize this below, with format weight $w_{\mathrm{fmt}}=0.2$ and length scale $\lambda_{\mathrm{len}}=0.75$, where $r_{task}$ can be $r_{cls}$ or $r_{qa}$, depending on if the sample belongs to a classification or QA task respectively:
\[
r=(1-w_{\mathrm{fmt}})\,r_{\mathrm{task}} + w_{\mathrm{fmt}}\,r_{\mathrm{fmt}} + \lambda_{\mathrm{len}}\,r_{\mathrm{len}},\]

We provide additional details on $r_{task}$, $r_{fmt}$, $r_{len}$. For $r_{task}$, it can either be classification reward $r_{cls}$, or question-answering reward $r_{qa}$, depending on whether the sample involves a classification response or a free-text answer respectively. $r_{cls}$ is computed using a binary score for whether the predicted label exactly matches the ground truth label. $r_{qa}$ leverages a cosine similarity reward that compares the embedding of the predicted free text sequence and the ground truth, leveraging MiniLM-L6-v2\footnote{https://huggingface.co/sentence-transformers/all-MiniLM-L6-v2} as the embedding model. Because cosine similarity typically falls between $[-1,1]$, we ensure that the values are transformed into the range of $[0,1]$, such that it is compatible with reward assignment. We sumamarise this below, where $\hat{y}$ is the predicted response (which contains just the answer without reasoning trace) and $y$ is the ground truth:

\[
r_{\text{task}}(\hat{y},y)=
\begin{cases}
r_{\text{cls}}(\hat{y},y), & \text{if classification},\\[2pt]
r_{\text{qa}}(\hat{y},y), & \text{if free-text QA}.
\end{cases}
\]

\[
r_{\text{cls}}(\hat{y},y)=
\begin{cases}
1, & \hat{y}=y,\\
0, & \text{otherwise}.
\end{cases}
\]

\[
r_{\text{qa}}(\hat{y},y)=\frac{\cos\!\left(e(\hat{y}),e(y)\right)+1}{2},
\qquad
\cos\!\left(u,v\right)=\frac{u^\top v}{\lVert u\rVert_2\,\lVert v\rVert_2},
\qquad
u=e(\hat{y}),\; v=e(y).
\]

For the formatting reward $r_{fmt}$, we assign a binary reward based on whether the generated response strictly follows the output format specified in the prompt. In particular, we provide the model the following system prompt:
\begin{lstlisting}[breaklines=true]
{content}
You FIRST think about the reasoning process as an internal monologue and then provide the final answer. The reasoning process MUST BE enclosed within <think> </think> tags. The final answer MUST BE put in \boxed{}.
\end{lstlisting}

A response receives $r_{fmt}$ = 1 if and only if it contains \texttt{<think>} and \texttt{</think>} tags, followed by a final prediction enclosed in \texttt{\textbackslash boxed{}}, in this order; otherwise, $r_{\mathrm{fmt}}$ = 0. This check is applied to the full response and enforces adherence to the prescribed output schema. 

Finally, we provide a overlong length penalty, $r_{len}$ which follows~\citet{zhang2025grpo_lead} to prevent excessive length and verbosity of responses. Accordingly, let $\ell$ be response length, we use a max $L_{\max}$ of 812 tokens and buffer length $B$ of 128 tokens. 
The penalty is:
\[
r_{\mathrm{len}}(\ell)=
\begin{cases}
0, & \ell \le L_{\max}-B,\\
\frac{(L_{\max}-B)-\ell}{B}, & L_{\max}-B < \ell \le L_{\max},\\
-1, & \ell > L_{\max}.
\end{cases}
\]

\section{Reasoning Evaluation Setup}
\label{app:reasoning_eval_setup}

We provide the full details of the setup for evaluating the reasoning trace interpretability in the following. Metrics are computed over the union of held-out zero-shot tasks (AUT, SER, IDR, SMSA, SIR), with all models evaluated under identical prompt templates, data loading, and multimodal preprocessing. Generation is capped at $T_{\max}=512$ new tokens per sample.

\textbf{Consistency.} For each sample $x$, $N=5$ independent reasoning traces $\{\hat{y}_i(x)\}_{i=1}^{N}$ are drawn with sampling parameters set to each model's recommended decoding configuration, where $T$ denotes the softmax temperature, $p$ the nucleus (top-$p$) cumulative probability threshold, and $k$ the top-$k$ truncation cutoff. We use $(T,\, p,\, k) = (0.6,\, 0.95,\, 20)$ for Qwen2.5-Omni-7B, HumanOmniV2, OmniSapiens-7B-RL, and \namesm, and $(T,\, p,\, k) = (1.0,\, 0.95,\, 64)$ for Gemma~4-E4B. Consistency is the fraction of instances on which $N$ independent reasoning traces produce the same answer for a given sample $x$, averaged across all samples:
\begin{equation}
\text{Consistency} \;=\; \frac{1}{|\mathcal{D}|} \sum_{x \in \mathcal{D}} \mathbb{1}\!\left[\, |\{\hat{y}_1(x), \ldots, \hat{y}_N(x)\}| = 1 \,\right],
\end{equation}

\textbf{Average tokens.} Average tokens are computed on the same $N$ traces by extracting the reasoning content between the model's thinking delimiters and tokenising with each model's own tokenizer, so that reported lengths reflect the actual inference cost incurred per model.

\textbf{Paraphrase.} A single reasoning trace is drawn under the same stochastic decoding configuration (as per consistency evaluation setup) on a semantically equivalent paraphrase $\tilde{x}$ of the original prompt, yielding $\hat{y}_{\text{para}}(x)$. Paraphrased prompts are pre-generated offline (one paraphrase per test instance), and held fixed across all models so that robustness is measured under the same input perturbation for every method; only the natural-language instruction is paraphrased, while task labels, output format, and multimodal inputs are preserved verbatim. Paraphrase consistency is the agreement between the original and paraphrased predictions, where $\hat{y}(x)$ denotes the first stochastic sample on the original prompt under the same decoding parameters, and we average this score over all samples. We provide the paraphrased prompts in the code release: \url{https://github.com/MIT-MI/human_behavior_atlas}.
\begin{equation}
\text{Paraphrase} \;=\; \frac{1}{|\mathcal{D}|} \sum_{x \in \mathcal{D}} \mathbb{1}\!\left[\, \hat{y}(x) = \hat{y}_{\text{para}}(x) \,\right],
\end{equation}

\textbf{Human evaluation.} Three annotators with backgrounds in interpretability, human–computer interaction, and healthcare, domains directly relevant to the critical applications of behavioral models, conducted pairwise comparisons between \namesm\ and each baseline. For each instance, annotators are shown two reasoning traces (A and B) extracted from the \texttt{<think>\ldots</think>} block of each model, with the task question provided as optional context, and select A, B, or Tie independently along three criteria mapped to XAI desiderata~\citep{alvarez2018towardsinterp}: specificity (completeness), coherence (understandability), and concision (compactness). For each baseline $b$ and criterion $c$, let $w_b^c$, $\ell_b^c$, $t_b^c$ denote the counts of \namesm-preferred, baseline-preferred, and tied judgments with $n_b^c = w_b^c + \ell_b^c + t_b^c$; we report the rates for $\text{Tie+Win} = (w_b^c + t_b^c)/n_b^c$, $\text{Tie} = t_b^c/n_b^c$, and $\text{Win}^\dagger = w_b^c/(w_b^c + \ell_b^c)$, where $\text{Win}^\dagger$ isolates the win rate excluding ties. Aggregate rates in Tab.~\ref{tab:human_eval} average across baselines. The annotation instructions shown in the box below. To ensure fair and robust evaluation, the annotators were not part of the authorship of this paper, and the \namesm\ and baselines are masked with letters A and B.

\begin{tcolorbox}[
    colback=gray!5,
    colframe=gray!50,
    title=\textbf{Annotation Instructions},
    fonttitle=\bfseries,
    boxrule=0.5pt,
    arc=2pt,
    left=6pt, right=6pt, top=4pt, bottom=4pt,
]
\small
You will be shown pairs of reasoning explanations (A and B) for the same prediction. The reasoning is between the \texttt{<think>}\ \texttt{</think>} tags. Your task is to compare them and select which explanation is better along each criterion. For each row, choose \textbf{A} if Explanation A is better, \textbf{B} if Explanation B is better, or \textbf{Tie} if they are similar. Place A, B, or Tie under the Specificity, Coherence, and Concision columns respectively.

\smallskip
\textbf{Evaluation Criteria.}
\begin{itemize}
    \setlength{\itemsep}{0pt}
    \item \textbf{Specificity:} Which explanation is more concrete and avoids vague or generic statements?
    \item \textbf{Coherence:} Which explanation is more logically structured and easier to follow?
    \item \textbf{Concision:} Which explanation conveys its key points more efficiently without unnecessary verbosity?
\end{itemize}

\textbf{Notes.} Focus only on the quality of the reasoning explanation. You may consult the task question for optional context. Evaluate based on the explanations themselves. Work quickly and rely on your first judgment.
\end{tcolorbox}

\section{Additional Formulas}

\subsection{Full GRPO Formulation}\label{app:grpo_full_formulation}
For completeness, we provide the full formulation of GRPO in Sec.~\ref{sec:prelim}, including the surrogate objective and importance-sampling formulation. 

Accordingly, for task $m$ and sample $q$, GRPO samples a rollout group $G_{(m,q)}$ of responses $\{o_{(m,q,i)}\}$, where $i \in G_{(m,q)}$ indexes individual rollouts (i.e., a sampled response) with rewards $r_{(m,q,i)}$, computing the group-normalized advantage:
\begin{equation}
\label{eq:grpo_adv2}
\hat A_{(m,q,i)}
=
\frac{r_{(m,q,i)}-\hat\mu_{G_{(m,q)}}}{\hat\sigma_{G_{(m,q)}}+\varepsilon},
\end{equation}
where $\hat\mu_{G_{(m,q)}}$ and $\hat\sigma_{G_{(m,q)}}$ are the empirical mean
and standard deviation of $\{r_{(m,q,i)}\}_{i=1}^{|G_{(m,q)}|}$. GRPO then optimizes $\pi_\theta(a\mid s)$ by performing a PPO-style trust-region update. At token position $k$ of response $o_{(m,q,i)}$,
$\varphi_{(m,q,i):k}(\theta)$ denotes the importance sampling ratio between
$\pi_\theta$ and the old policy $\pi_{\theta_{\mathrm{old}}}$,
$\tilde A_{(m,q,i):k}(\theta)$ denotes the PPO-clipped surrogate using
$\hat A_{(m,q,i)}$, and $J_{\mathrm{GRPO}}(\theta)$ averages this surrogate over
tokens and rollout samples with an optional KL penalty to a reference policy
$\pi_{\mathrm{ref}}$ (with weight $\beta$). We summarize these with a compact
objective:
\begin{equation}
\label{eq:grpo_compact2}
\begin{aligned}
\varphi_{(m,q,i):k}(\theta)
&=
\frac{
\pi_\theta\!\big(o_{(m,q,i):k}\mid q,\, o_{(m,q,i):<k}\big)
}{
\pi_{\theta_{\mathrm{old}}}\!\big(o_{(m,q,i):k}\mid q,\, o_{(m,q,i):<k}\big)
}
\\
\tilde A_{(m,q,i):k}(\theta)
&=
\min\!\Big(
\varphi_{(m,q,i):k}(\theta)\,\hat A_{(m,q,i)},
\\
& \hspace{-10pt} \phantom{=\min\!\Big(}
\operatorname{clip}\!\big(\varphi_{(m,q,i):k}(\theta),\,1-\epsilon,\,1+\epsilon\big)\,\hat A_{(m,q,i)}
\Big)
\\
J_{\mathrm{GRPO}}(\theta)
&=
\mathbb{E}_{q\sim\mathcal D_m}
\mathbb{E}_{\{o_{(m,q,i)}\}\sim \pi_{\theta_{\mathrm{old}}}}\!\Bigg[
\frac{1}{|G_{(m,q)}|}
\sum_{i=1}^{|G_{(m,q)}|}
\\
&\hspace{-60pt}\phantom{=\mathbb{E}\Bigg[}
\frac{1}{n_{o_{(m,q,i)}}}
\sum_{k=1}^{n_{o_{(m,q,i)}}}
\tilde A_{(m,q,i):k}(\theta)
\Bigg]
\;-\;
\beta\,
\mathbb{E}\!\left[
D_{\mathrm{KL}}\!\left(\pi_\theta\;\|\;\pi_{\mathrm{ref}}\right)
\right].
\end{aligned}
\end{equation}

\subsection{Additional HARPO Details}~\label{app:harpo_full_formulation}

In Sec.~\ref{sec:harpo}, we explained the structured modulation utilized by HARPO, which constructs modulation factors by comparing contribution signals to a geometric mean reference. We provide further details on why this construction yields modulation factors whose geometric mean equals 1, ensuring that multiplicative upscaling from certain modulation factors is exactly compensated by downscaling from others.

Fix an iteration $t$ and a task $m$, and let $\mathcal{Q}_m^{(t)}$ denote the set of samples associated with task $m$. The sample-level geometric reference is defined as:
\[
\bar p_{\mathrm{ref},m}^{(t)}
=
\Bigg(
\prod_{q \in \mathcal{Q}_m^{(t)}} p_{(m,q)}^{(t)}
\Bigg)^{\frac{1}{|\mathcal{Q}_m^{(t)}|}}.
\]

Using this reference, the sample-level modulation factor for each sample $q$ is constructed as:
\[
s_{(m,q)}^{(t)} = \frac{\bar p_{\mathrm{ref},m}^{(t)}}{p_{(m,q)}^{(t)}}.
\]

Taking the product over all samples in $\mathcal{Q}_m^{(t)}$ yields:
\[
\prod_{q \in \mathcal{Q}_m^{(t)}} s_{(m,q)}^{(t)}
=
\prod_{q \in \mathcal{Q}_m^{(t)}} \frac{\bar p_{\mathrm{ref},m}^{(t)}}{p_{(m,q)}^{(t)}}
=
\frac{(\bar p_{\mathrm{ref},m}^{(t)})^{|\mathcal{Q}_m^{(t)}|}}
{\prod_{q \in \mathcal{Q}_m^{(t)}} p_{(m,q)}^{(t)}}.
\]

By definition of the geometric mean:
\[
(\bar p_{\mathrm{ref},m}^{(t)})^{|\mathcal{Q}_m^{(t)}|}
=
\prod_{q \in \mathcal{Q}_m^{(t)}} p_{(m,q)}^{(t)},
\]
and therefore:
\[
\prod_{q \in \mathcal{Q}_m^{(t)}} s_{(m,q)}^{(t)} = 1.
\]

This shows that the sample-level modulation factors possess a geometric mean of 1.

An identical argument applies at the task-level. Let $\mathcal{M}$ denote the set of tasks and define the task-level geometric reference as
\[
\bar p_{\mathrm{ref},\mathcal{M}}^{(t)}
=
\Bigg(
\prod_{m \in \mathcal{M}} p_m^{(t)}
\Bigg)^{\frac{1}{|\mathcal{M}|}}.
\]

The task-level modulation factors are given by
\[
s_m^{(t)} = \frac{\bar p_{\mathrm{ref},\mathcal{M}}^{(t)}}{p_m^{(t)}}.
\]

Taking the product over all tasks yields
\[
\prod_{m \in \mathcal{M}} s_m^{(t)} = 1.
\]

Therefore, the modulation factors are possess a geometric mean of 1 at both the sample and task-levels, ensuring that multiplicative upscaling from some modulation factors is exactly compensated by downscaling from others. Hence, the factors cannot simulateneously enlarge or shrink all advantages at the sample or task-level, mitigating unintended influence on the effective global step size.

\clearpage
\section{Additional Results}~\label{app:additional_results}
\vspace{-20pt}

\subsection{Full Ablation Results}~\label{app:full_ablation_results}
\vspace{-20pt}

We provide the per task and per dataset breakdown of the results from the ablation in the following. Tab.~\ref{tab:harpo_ablations_full} represents the task-level performance of the ablations, whereas Tab.~\ref{tab:harpo_ablation_full_dataset_results} represents the per-dataset performance breakdown of all ablations.

\begin{table*}[h]
\vspace{-15pt}
\small
\centering
\caption{
Per-task performance (\%) for \names\ ablations across behavioral tasks.
Each value is the arithmetic mean over datasets associated with the task.
Avg. Rank is computed across tasks using per-task ranks (higher is better; ties use average rank), and then averaged over tasks (lower is better).
Rows with missing task values are omitted from ranking.
}
\setlength{\tabcolsep}{7pt}
\renewcommand{\arraystretch}{0.95}
\resizebox{\linewidth}{!}{%
\begin{tabular}{lccccccccccc}
\toprule
\textbf{Variant}
& \textbf{EMO}
& \textbf{HUM}
& \textbf{INT}
& \textbf{PTSD}
& \textbf{ANX}
& \textbf{DEP}
& \textbf{SEN}
& \textbf{SAR}
& \textbf{SOC}
& \textbf{NVC}
& \textbf{Avg. Rank} $\downarrow$ \\
\midrule
\textbf{\names}
& 76.55 
& \textbf{69.85} 
& 50.52 
& \textbf{98.39} 
& \textbf{91.98} 
& \textbf{78.87} 
& \textbf{77.61} 
& \underline{70.64} 
& 25.40 
& \underline{14.54} 
& \textbf{1.90} \\
w/o structured modulation
& \textbf{78.44} 
& 66.61 
& 50.38 
& \textbf{98.39} 
& \underline{90.68} 
& \underline{77.01} 
& \underline{77.48} 
& \textbf{72.68} 
& 29.08 
& \textbf{15.12} 
& \underline{2.00} \\
w/o inertial control
& 76.42 
& 63.05 
& \textbf{53.28} 
& \textbf{98.39} 
& \underline{90.68} 
& 76.48 
& 71.92 
& 62.68 
& \textbf{30.12} 
& 13.95 
& 2.70 \\
w/o sample-level modulation
& \underline{77.27} 
& \underline{67.82} 
& \underline{50.94} 
& \textbf{98.39} 
& \underline{90.68} 
& 71.04 
& 76.37 
& 68.19 
& \underline{29.43} 
& 12.21 
& 2.60 \\
\bottomrule
\end{tabular}
}
\label{tab:harpo_ablations_full}
\end{table*}

\begin{table*}[h]
\small
\centering
\vspace{-15pt}
\caption{Ablation results for HARPO components, grouped by behavioral tasks and datasets. Following the unified metrics proposed in the Human Behavior Atlas Benchmark~\citep{ong2025human}, we use binary weighted F1 for SEN; mean per-class weighted accuracy for EMO; weighted F1 for HUM, SAR, ANX, DEP, PTSD; and LLM-Judge accuracy for SOC, INT, NVC.}
\renewcommand{\arraystretch}{1}
\setlength{\tabcolsep}{2pt}
\resizebox{\linewidth}{!}{%
\begin{tabular}{p{4.8cm}ccccccccccccccccc}
\toprule
\textbf{Model / Ablations} 
& \multicolumn{4}{c}{\textbf{EMO}} 
& \textbf{HUM}
& \textbf{INT}
& \textbf{PTSD} 
& \textbf{ANX}
& \multicolumn{2}{c}{\textbf{DEP}} 
& \multicolumn{3}{c}{\textbf{SEN}} 
& \textbf{SAR} 
& \textbf{SOC} 
& \textbf{NVC} \\
\cmidrule(lr){2-5} \cmidrule(lr){6-6} \cmidrule(lr){7-7} \cmidrule(lr){8-8} 
\cmidrule(lr){9-9} \cmidrule(lr){10-11} \cmidrule(lr){12-14} 
\cmidrule(lr){15-15} \cmidrule(lr){16-16} \cmidrule(lr){17-17}
& \rotatebox{90}{CREMA-D} 
& \rotatebox{90}{MELD (E)} 
& \rotatebox{90}{MOSEI (E)} 
& \rotatebox{90}{TESS}
& \rotatebox{90}{UR-FUNNY}
& \rotatebox{90}{IntentQA}
& \rotatebox{90}{PTSD\_WILD}
& \rotatebox{90}{MMPSY (A)}
& \rotatebox{90}{MMPSY (D)} 
& \rotatebox{90}{DAIC--WOZ}
& \rotatebox{90}{MELD (S)} 
& \rotatebox{90}{CH-SIMSv2} 
& \rotatebox{90}{MOSEI (S)}
& \rotatebox{90}{MUStARD}
& \rotatebox{90}{Social-IQ 2.0}
& \rotatebox{90}{MimeQA} \\
\midrule\midrule

\textbf{HARPO}
& 85.80 & 69.14 & 55.45 & 95.83
& 69.85 & 50.52 & 98.39 & 91.98
& 84.53 & 73.20
& 75.49 & 88.71 & 68.64
& 70.64 & 25.40 & 14.54 \\

w/o structured modulation
& 84.48 & 71.70 & 60.36 & 97.22
& 66.61 & 50.38 & 98.39 & 90.68
& 83.20 & 70.82
& 70.43 & 88.37 & 73.65
& 72.68 & 29.08 & 15.12 \\

w/o inertial control
& 86.49 & 67.86 & 56.61 & 94.72
& 63.05 & 53.28 & 98.39 & 90.68
& 82.04 & 70.92
& 73.80 & 84.65 & 57.32
& 62.68 & 30.12 & 13.95 \\

w/o sample-level modulation
& 84.71 & 68.92 & 57.38 & 98.06
& 67.82 & 50.94 & 98.39 & 90.68
& 73.44 & 68.64
& 71.61 & 87.76 & 69.74
& 68.19 & 29.43 & 12.21 \\

\bottomrule
\end{tabular}
}
\label{tab:harpo_ablation_full_dataset_results}
\vspace{-10pt}
\end{table*}

\subsection{Full Dataset Results}
For brevity, Tab.~\ref{tab:models_and_algos_multi_task_results} shows the performance at the task-level. We additionally provide the dataset-level breakdown of these results in the following Tab.~\ref{tab:full_dataset_results}. We also provided the full dataset breakdown for HARPO and its ablations in Tab.~\ref{tab:harpo_ablation_full_dataset_results}.

\begin{table*}[h]
\small
\centering
\caption{Full results grouped by behavioral tasks (headers) and their relevant datasets (sub-headers). Following the unified metrics proposed in the Human Behavior Atlas Benchmark~\citep{ong2025human}, we use binary weighted F1 for SEN; mean per-class weighted accuracy for EMO; weighted F1 for HUM, SAR, ANX, DEP, PTSD; and LLM-Judge accuracy for SOC, INT, NVC.}
\renewcommand{\arraystretch}{1}
\setlength{\tabcolsep}{2pt}
\resizebox{\linewidth}{!}{%
\begin{tabular}{p{4.8cm}ccccccccccccccccc}
\toprule
\textbf{Model / Algorithm} 
& \multicolumn{4}{c}{\textbf{EMO}} 
& \textbf{HUM}
& \textbf{INT}
& \textbf{PTSD} 
& \textbf{ANX}
& \multicolumn{2}{c}{\textbf{DEP}} 
& \multicolumn{3}{c}{\textbf{SEN}} 
& \textbf{SAR} 
& \textbf{SOC} 
& \textbf{NVC} \\
\cmidrule(lr){2-5} \cmidrule(lr){6-6} \cmidrule(lr){7-7} \cmidrule(lr){8-8} 
\cmidrule(lr){9-9} \cmidrule(lr){10-11} \cmidrule(lr){12-14} 
\cmidrule(lr){15-15} \cmidrule(lr){16-16} \cmidrule(lr){17-17}
& \rotatebox{90}{CREMA-D} 
& \rotatebox{90}{MELD (E)} 
& \rotatebox{90}{MOSEI (E)} 
& \rotatebox{90}{TESS}
& \rotatebox{90}{UR-FUNNY}
& \rotatebox{90}{IntentQA}
& \rotatebox{90}{PTSD\_WILD}
& \rotatebox{90}{MMPSY (A)}
& \rotatebox{90}{MMPSY (D)} 
& \rotatebox{90}{DAIC--WOZ}
& \rotatebox{90}{MELD (S)} 
& \rotatebox{90}{CH-SIMSv2} 
& \rotatebox{90}{MOSEI (S)}
& \rotatebox{90}{MUStARD}
& \rotatebox{90}{Social-IQ 2.0}
& \rotatebox{90}{MimeQA} \\
\midrule\midrule
\multicolumn{17}{l}{\textbf{Models}} \\
\midrule
Gemma-4-E4B (8B)
& 50.21 & 66.02 & 58.63 & 49.07
& 52.63 & 40.25 & 86.00 & 33.36
& 16.05 & 29.05
& 68.89 & 76.53 & 63.24
& 73.54 & 20.40 & 4.65 \\
Gemma-3-4B
& 49.50 & 64.20 & 56.50 & 49.90
& 59.70 & 22.70 & 49.90 & 60.10
& 78.80 & 13.70
& 78.50 & 81.30 & 61.70
& 52.90 & 19.10 & 2.30 \\
Qwen 2.5-Omni-7B
& 52.10 & 66.10 & 58.00 & 56.80
& 54.30 & 25.40 & 76.00 & 79.30
& 79.10 & 63.60
& 70.00 & 71.40 & 60.20
& 65.60 & 25.40 & 6.90 \\
Qwen-2.5-VL-7B
& 50.10 & 57.10 & 59.20 & 49.90
& 58.30 & 24.90 & 75.50 & 63.10
& 65.30 & 62.30
& 67.40 & 52.40 & 31.70
& 51.10 & 23.10 & 9.80 \\
Qwen 3-VL-8B-Instruct
& 50.20 & 70.87 & 59.44 & 50.14
& 66.76 & 38.00 & 92.70 & 42.29
& 36.38 & 66.86
& 68.97 & 77.19 & 62.93
& 63.67 & 24.94 & 13.95 \\
\textsc{OmniSapiens-7B RL}
& 50.10 & 69.90 & 58.10 & 51.00
& 63.90 & 48.60 & 96.80 & 91.90
& 81.40 & 72.90
& 57.10 & 39.30 & 22.40
& 64.70 & 30.40 & 13.30 \\
HumanOmniV2-7B
& 56.00 & 63.30 & 55.80 & 63.70
& 63.80 & 26.30 & 82.40 & 52.70
& 67.20 & 63.60
& 76.80 & 82.50 & 63.30
& 39.50 & 28.20 & 9.30 \\
\midrule
\textbf{\namesm\ (ours)}
& 85.80 & 69.14 & 55.45 & 95.83
& 69.85 & 50.52 & 98.39 & 91.98
& 84.53 & 73.20
& 75.49 & 88.71 & 68.64
& 70.64 & 25.40 & 14.54 \\
\midrule\midrule
\multicolumn{17}{l}{\textbf{Training Algorithms}} \\
\midrule
RLOO
& 85.77 & 67.74 & 50.05 & 98.75
& 67.86 & 51.73 & 98.39 & 90.68
& 84.26 & 70.88
& 67.72 & 89.64 & 73.21
& 62.58 & 29.54 & 16.28 \\
RE++
& 82.56 & 66.52 & 59.05 & 95.56
& 60.26 & 5.01 & 98.39 & 93.11
& 79.09 & 68.64
& 66.67 & 87.86 & 15.03
& 50.21 & 12.64 & 4.07 \\
GPG
& 82.62 & 69.36 & 60.00 & 98.75
& 69.28 & 54.21 & 98.39 & 90.36
& 85.88 & 70.92
& 75.10 & 87.77 & 64.43
& 45.96 & 27.93 & 12.79 \\
GRPO
& 82.88 & 69.85 & 57.22 & 95.83
& 27.56 & 49.90 & 98.39 & 90.42
& 84.35 & 70.92
& 76.28 & 84.57 & 71.68
& 53.58 & 23.30 & 11.00 \\
EMA-GRPO
& 76.21 & 65.37 & 56.85 & 95.00
& 63.50 & 52.62 & 98.39 & 90.68
& 75.84 & 70.88
& 65.94 & 76.94 & 61.96
& 77.75 & 30.58 & 15.70 \\
\midrule
\textbf{\names\ (ours)}
& 85.80 & 69.14 & 55.45 & 95.83
& 69.85 & 50.52 & 98.39 & 91.98
& 84.53 & 73.20
& 75.49 & 88.71 & 68.64
& 70.64 & 25.40 & 14.54 \\
\bottomrule
\end{tabular}
}
\label{tab:full_dataset_results}
\end{table*}

\subsection{Additional Training Plots}~\label{app:additional_plots} 

We provide additional training plots to empirically illustrate the training dynamics in our experiments. 

\begin{figure}[h!]
    \centering
    \vspace{-50pt}
    \includegraphics[width=\linewidth]{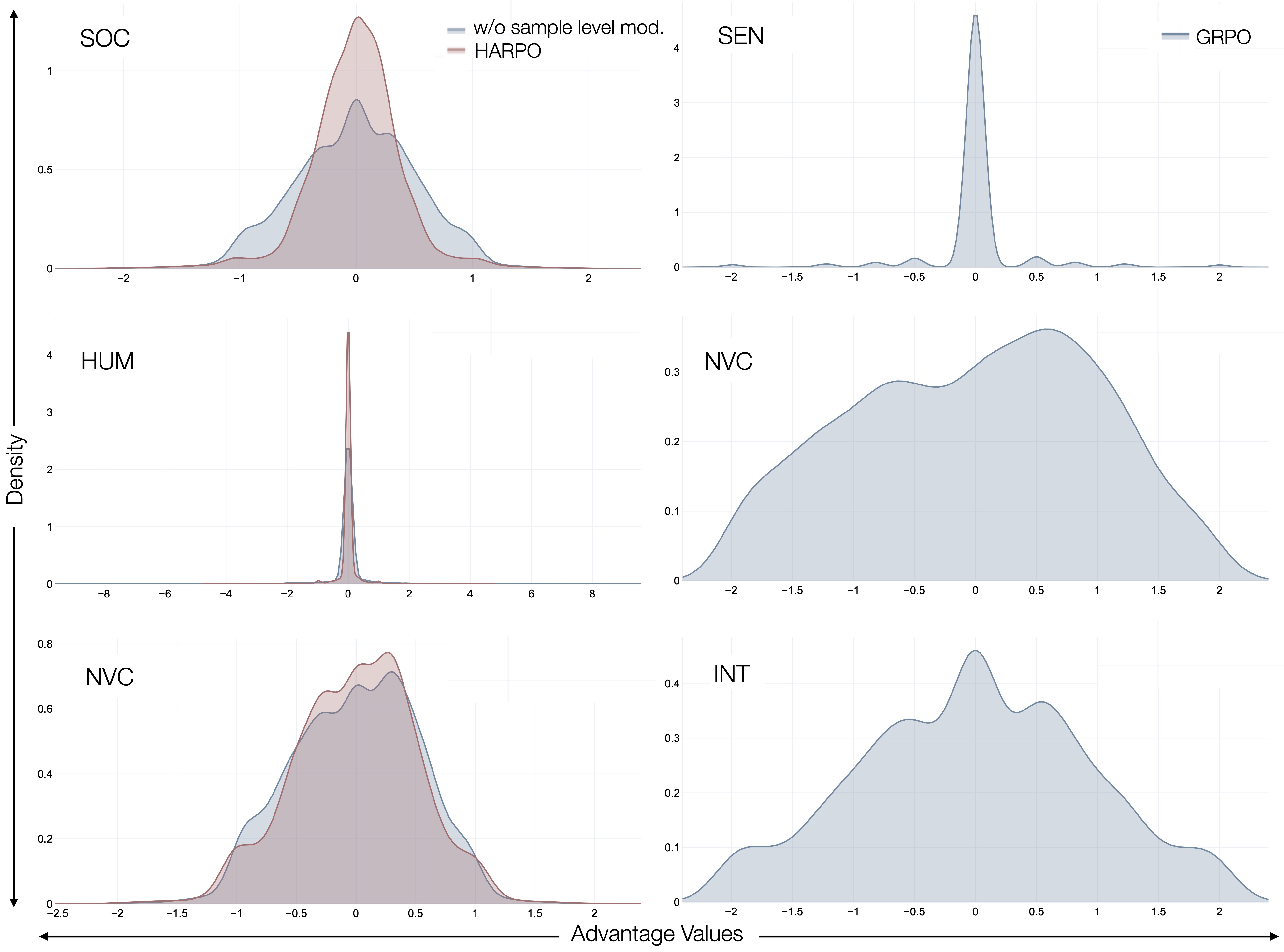}
    \caption{Left: Advantage distributions for HARPO (red) and the ablation (blue) which does not involve sample-level modulation, across the different tasks SOC, HUM, NVC. We observe that HARPO, with sample-level modulation induces a narrower advantage distribution. Right: More plots to depict the group-normalized advantage distributions of different tasks, SEN, NVC, INT. We observe that the different behavioral tasks provide different advantage distributions.
    }
    \label{fig:appendix_adv_dist}
\end{figure}

We include a Fig.~\ref{fig:appendix_adv_dist} that depicts the advantage distributions of the different behavioral tasks in the Human Behavior Atlas benchmark~\citep{ong2025human}. In particular, the right column of Fig.~\ref{fig:appendix_adv_dist} depicts the group-normalized advantages of the tasks SEN, NVC, INT under a GRPO run. The different advantage distributions observed highlights how the behavioral tasks can systematically induce different advantages, as mentioned in Sec.~\ref{sec:prelim}. 

In the left column of this Fig.~\ref{fig:appendix_adv_dist}, we compare the advantage distributions with and without sample-level modulation for other tasks (SOC, HUM, NVC). Accordingly, we observe that with sample-level modulation in the HARPO method, the advantage distributions tend to become narrower than without. This coincides with more consistent performance across tasks in Tab.~\ref{tab:harpo_ablations} with HARPO achieving an average task rank of 1.90 compared to without sample-level modulation at 2.60.

\begin{figure}[h]
    \centering
    \includegraphics[width=\linewidth]{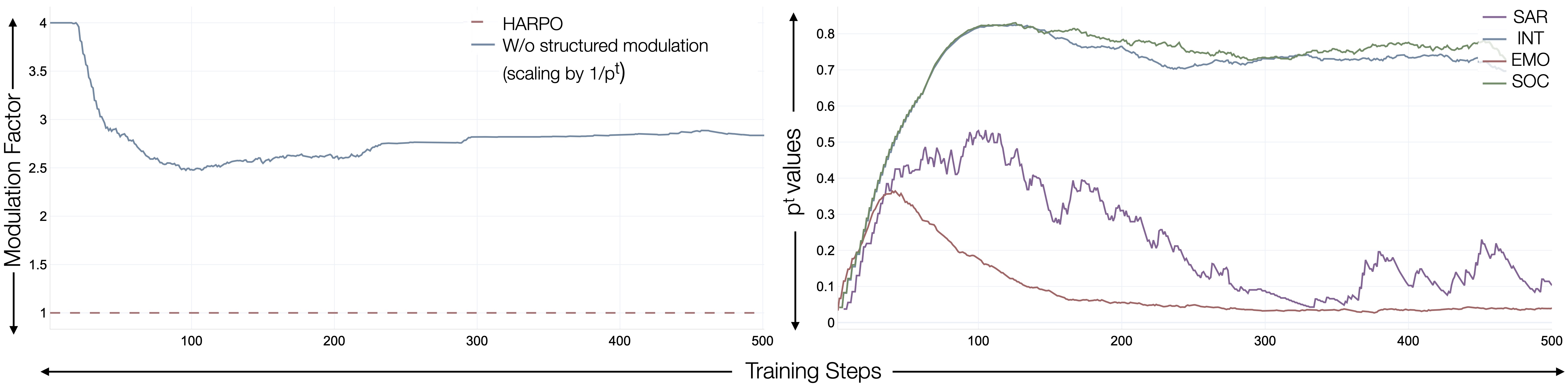}
    \caption{Left: Comparison of geometric mean of the task-level modulation factors $s^{(t)}_m$ between HARPO and the w/o structured modulation ablation that utilizes $s^{(t)}=1 / p^{(t)}$. We observe that the modulation geometric mean of the modulation factors are above 2.5 throughtout training. Right: Values of $p^{(t)}$ over training for different tasks SAR, INT, EMO, SOC. We observe that between specific tasks (i.e., INT vs EMO), the task-level contribution signals $p^{(t)}_{m}$ can vary by considerable orders of magnitude throughout training.}
    \label{fig:appendix_add_factors}
\end{figure}

We include a Fig.~\ref{fig:appendix_add_factors} to illustrate the task-level modulation factors over time between the ablation that utilizes $s^{(t)}=1 / p^{(t)}$ and HARPO. To this end, we observe that in the left column of this figure, the geometric mean of the modulation factors are consistently above 2.5. This further illustrates how the modulation factors can induce unintended scaling of the global step size, without geometric centering.

In the same Fig.~\ref{fig:appendix_add_factors}, in the right column, we also observe how the contribution signals $p^{(t)}$ varies with training for different tasks, SAR, INT, EMO, SOC. Accordingly, we observe that for certain tasks, the task-level contribution signals $p^{(t)}_{m}$ can differ by orders of magnitude. For example, the $p^{(t)}_{m}$ for INT may exceed EMO by a factor of up to 7. Since these signals are used to construct the modulation factors, the empirical illustrations provide additional context for the considerable variation in modulation factors. This motivates the use of a geometric reference as a practical tool to temper excessively large differences in modulation factors.

\end{document}